\documentclass[pdflatex,sn-basic,Numbered]{sn-jnl}

\usepackage{graphicx}
\usepackage{multirow}
\usepackage{amsmath,amssymb,amsfonts,bm}
\usepackage{booktabs}
\usepackage[table]{xcolor}
\usepackage{tabularx}
\usepackage{ragged2e}
\usepackage{algorithm}
\usepackage{algorithmic}
\usepackage{float}
\usepackage{placeins}
\usepackage{flafter}
\usepackage{listings}
\usepackage{url}
\usepackage{caption}
\usepackage{tikz}
\usetikzlibrary{arrows.meta,positioning,fit,calc}

\definecolor{oursrow}{RGB}{235,245,255}
\newcommand{\best}[1]{\textbf{#1}}
\newcommand{\second}[1]{\underline{#1}}

\begin{document}

\title[UGOD for Sparse-View 3DGS]{UGOD: Uncertainty-Guided Opacity and Dropout for Sparse-View 3D Gaussian Splatting}

\author[1]{\fnm{Zhihao} \sur{Guo}}
\author*[2]{\fnm{Peng} \sur{Wang}}\email{pw0038@surrey.ac.uk}
\author[3]{\fnm{Zidong} \sur{Chen}}
\author[4]{\fnm{Xiangyu} \sur{Kong}}
\author[5]{\fnm{Yan} \sur{Lyu}}
\author[6]{\fnm{Guanyu} \sur{Gao}}
\author[7]{\fnm{Chenghao} \sur{Qian}}
\author[8]{\fnm{Ziyang} \sur{Wang}}
\author[1]{\fnm{Xinqi} \sur{Fan}}
\author[1]{\fnm{Liangxiu} \sur{Han}}

\affil[1]{\orgname{Manchester Metropolitan University}, \orgaddress{\city{Manchester}, \country{United Kingdom}}}
\affil*[2]{\orgname{University of Surrey}, \orgaddress{\city{Guildford}, \country{United Kingdom}}}
\affil[3]{\orgname{Imperial College London}, \orgaddress{\city{London}, \country{United Kingdom}}}
\affil[4]{\orgname{University of Exeter}, \orgaddress{\city{Exeter}, \country{United Kingdom}}}
\affil[5]{\orgname{Southeast University}, \orgaddress{\city{Nanjing}, \country{China}}}
\affil[6]{\orgname{Nanjing University of Science and Technology}, \orgaddress{\city{Nanjing}, \country{China}}}
\affil[7]{\orgname{University of Leeds}, \orgaddress{\city{Leeds}, \country{United Kingdom}}}
\affil[8]{\orgname{Aston University}, \orgaddress{\city{Birmingham}, \country{United Kingdom}}}

\abstract{Sparse-view 3D Gaussian Splatting is prone to overfitting because limited observations leave many Gaussian primitives weakly constrained, yet their contributions are still accumulated through alpha blending. Without uncertainty estimation, the renderer cannot distinguish unreliable primitives from well-constrained ones, allowing their erroneous contributions to corrupt novel-view synthesis. We introduce \textbf{UGOD}, an uncertainty-guided framework that estimates a view-dependent uncertainty score for each Gaussian and uses it to regulate its rendering contribution. A lightweight uncertainty head conditioned on Gaussian attributes and viewing direction predicts this score, which then drives a differentiable opacity-modulation mechanism that attenuates high-uncertainty primitives before compositing. During training, a detached soft-dropout branch applies an uncertainty-controlled continuous keep mask to discourage the model from relying on poorly constrained Gaussians and thereby reduce overfitting. Crucially, detaching the uncertainty score prevents gradients from this stochastic regulariser from biasing or collapsing the uncertainty prediction. Experiments on Mip-NeRF~360 and LLFF show that UGOD improves sparse-view novel-view synthesis while producing more compact Gaussian representations than the compared methods. These results demonstrate that Gaussian uncertainty provides an effective rendering-time control for sparse-view reconstruction.}


\maketitle

\begin{figure}[t]
  \centering
  \includegraphics[width=\textwidth]{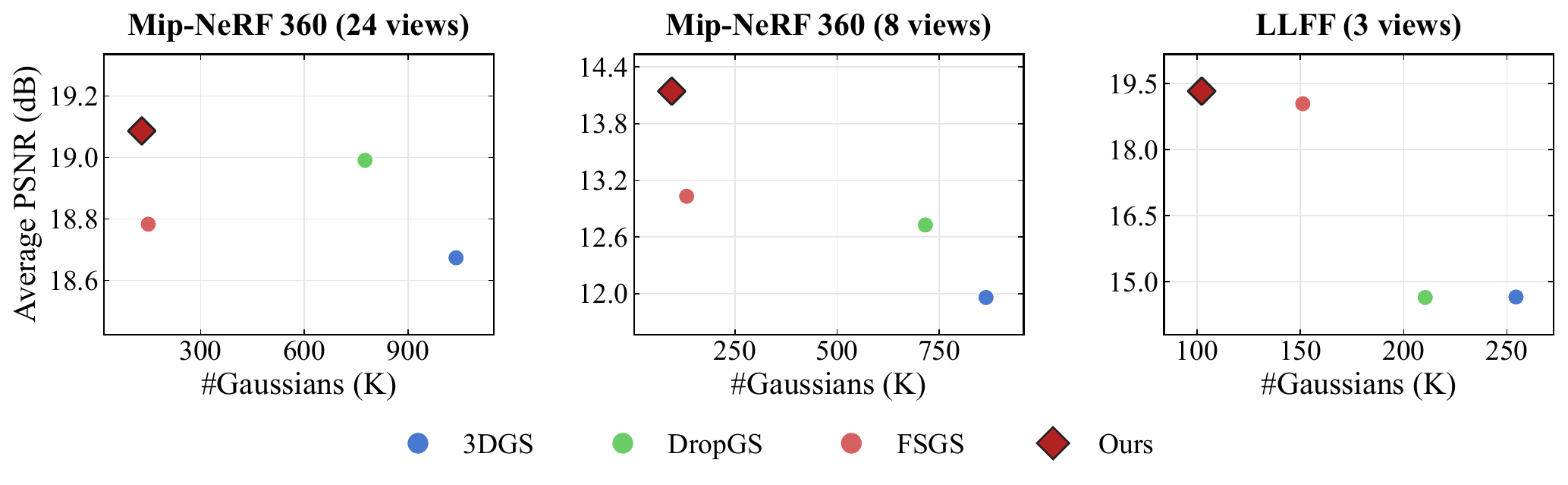}
  \caption{Average PSNR and final-Gaussian reduction versus 3DGS* across
  MiP-NeRF~360 (24/8 views) and LLFF (3 views). Higher is better on every
  axis; UGOD leads all six.}
  \label{fig:tradeoff_overview}
\end{figure}

\section{Introduction}
\label{introduction}
Novel-view synthesis (NVS) from a limited collection of images underpins
applications such as digital twinning, augmented and virtual reality, and
embodied robotics. 3D Gaussian Splatting (3DGS)~\cite{kerbl20233d} has become
an attractive representation for this task because it models a scene as
anisotropic Gaussian primitives that are projected, depth sorted, and alpha
blended by a rasteriser. This design avoids the dense ray sampling required by
many neural volumetric methods and enables high-quality real-time rendering.
Its effectiveness nevertheless depends on the coverage, geometric accuracy,
and appearance diversity provided by the input views.

These requirements are difficult to satisfy in sparse-view reconstruction.
Structure-from-Motion (SfM) initialisation may contain only a limited set of
points, and those points can be incomplete or inaccurate in weakly textured,
occluded, or poorly observed regions. The resulting Gaussians receive
insufficient multi-view supervision while their geometry and appearance are
optimised to explain the available images. Critically, the failure is not only
that geometry is incomplete: once a primitive is instantiated, its learned
opacity still determines how strongly it participates in alpha blending. An
unreliable Gaussian can therefore retain a large compositing weight, inject
incorrect colour into the accumulated ray, and reinforce a training-view
explanation that does not generalise. In other words, sparse-view artefacts
arise when weakly supported contributions are repeatedly accumulated by the
renderer.

Existing sparse-view 3DGS methods address this setting from complementary but
incomplete angles. Depth or other priors strengthen geometric guidance when
image evidence is scarce~\cite{xiong2025sparsegs,zhu2024fsgs}. Densification
increases point coverage when SfM initialisation is too sparse~\cite{zhu2024fsgs}.
Dropout- and pruning-based regularisers reduce the influence of selected
primitives during optimisation~\cite{park2025dropgaussian,zhang2024cor}. These
strategies improve reconstruction by adding constraints, growing the
representation, or filtering primitives in training. They do not, however,
explicitly ask whether a Gaussian is uncertain for the \emph{current} viewing
direction at the moment of alpha compositing, nor do they provide a
differentiable mechanism that attenuates its opacity before blending.

This gap motivates a rendering-centric view of sparse-view overfitting.
Opacity in conventional 3DGS is a view-agnostic scalar jointly optimised with
geometry and appearance; it does not separate a reliable contribution from one
that is poorly supported under the present camera. Because alpha blending is
order-dependent and cumulative, even a modest number of over-confident
primitives can dominate transmittance and colour along a ray. Regulating
opacity with a view-dependent uncertainty signal therefore targets the failure
mode directly: unreliable contributions should be down-weighted before they
enter compositing, rather than corrected only indirectly through denser
geometry or training-time dropping.

We introduce \textbf{UGOD}, an uncertainty-guided framework for sparse-view
3DGS. UGOD employs a lightweight per-Gaussian uncertainty head conditioned on
Gaussian attributes, appearance features, and viewing direction to estimate
view-dependent uncertainty. The normalised uncertainty drives a differentiable
opacity-modulation function that attenuates the learned opacity of
high-uncertainty primitives before compositing. During training, a detached
soft-dropout branch converts the same uncertainty signal into a continuous keep
mask, further regularising ambiguous Gaussians without allowing this stochastic
branch to distort the uncertainty prediction.

Across Mip-NeRF~360 and LLFF sparse-view protocols, UGOD improves novel-view
synthesis quality while producing compact Gaussian representations. These
results indicate that regulating compositing with view-dependent uncertainty
is an effective strategy for sparse-view 3DGS. The contributions of this work are:
\begin{itemize}
    \item A view-dependent, per-Gaussian uncertainty prediction that combines
    Gaussian attributes, appearance features, and viewing direction.
    \item A differentiable opacity gate that attenuates high-uncertainty
    primitives before alpha compositing.
    \item A gradient-detached soft-dropout regulariser that uses the learned
    uncertainty to stochastically regularise high-uncertainty primitives during
    training without directly updating the uncertainty estimator.
    \item Extensive sparse-view evaluation demonstrating improved novel-view
    synthesis quality and Gaussian compactness relative to the compared
    methods.
\end{itemize}
\section{Related Work}

\subsection{3D Gaussian Splatting}
3D Gaussian Splatting (3DGS) represents a scene using anisotropic Gaussian
primitives that are projected, depth sorted, and alpha blended in image space.
Its rasterisation-based renderer provides an attractive balance between
rendering quality and speed~\cite{kerbl20233d}. Subsequent studies have
improved camera handling~\cite{yu2024mip}, point management, and rendering
quality~\cite{yang2024gaussian,zhang2024pixel,bulo2024revising}. These advances
make 3DGS effective for dense-view reconstruction, but the quality of the
learned representation remains sensitive to the coverage and accuracy of its
initialisation and observations.

Opacity is optimised jointly with geometry and appearance, and determines each
primitive's contribution to alpha compositing. Existing studies have examined
the distinction between opacity and extinction formulations~\cite{celarek2025does},
volumetrically consistent Gaussian rasterisation~\cite{talegaonkar2024volumetrically},
and material-aware opacity modelling~\cite{yong2025omg}. These directions
improve the physical interpretation or material dependence of the rendering
weights. They do not, however, explicitly ask whether a Gaussian is reliable
for a particular viewing direction when the scene is weakly constrained. Our
work instead models this view-dependent uncertainty and uses it to regulate
opacity before alpha compositing.

\subsection{Sparse 3DGS Reconstruction}
\label{Sparse 3D Reconstruction}
Sparse-view reconstruction has been addressed through complementary forms of
geometric guidance and regularisation. Prior-based methods use additional
signals to compensate for incomplete image observations: SparseGS~\cite{xiong2025sparsegs},
for example, combines depth and diffusion priors. A second line of work seeks
to improve the representation before or during optimisation. FSGS~\cite{zhu2024fsgs}
densifies the initial point cloud to increase geometric coverage, which is
particularly helpful when SfM produces only a small or unevenly distributed
set of initial points. PcdGS~\cite{zhao2025pcdgs} similarly densifies sparse
SfM initialisation using mask and monocular-depth cues, whereas
LiDAR-3DGS~\cite{lim2025lidar3dgs} supplements it with an aligned LiDAR point
cloud when an additional sensor is available. These approaches improve the
geometric support available at initialisation.

Other approaches act directly on the Gaussian primitives. DropGaussian~\cite{park2025dropgaussian}
regularises optimisation by selectively dropping Gaussians, allowing the
remaining primitives to receive more informative gradients. CoR-GS~\cite{zhang2024cor}
uses geometric and rendering disagreement to guide co-pruning and pseudo-view
co-regularisation. SuraGS~\cite{shen2025surags} uses surface-aware primitives
for few-shot synthesis, while HR-2DGS~\cite{tang2025hr2dgs} combines depth and
normal regularisation in a 2DGS representation. StruGS~\cite{pang2025strugs}
instead targets structural consistency through multi-view guidance and
depth-balanced optimisation. Together, these methods show that sparse-view performance
depends not only on improving initial geometry but also on controlling which
primitives are retained and how they contribute during learning. Their
regularisation criteria, however, do not provide a view-dependent estimate of
each primitive's uncertainty at the point of rendering. Consequently,
the renderer cannot distinguish well-supported primitives from those
unreliable in the current view: both are assigned comparable importance during
compositing. Under cumulative alpha blending, an unreliable primitive can then
incorrectly consume transmittance or occlude reliable evidence, producing
view-dependent artefacts in the rendered image.

\subsection{Uncertainty Estimation in Gaussian Splatting}
Recent uncertainty-estimation methods for Gaussian splatting pursue objectives
that differ from sparse-view reconstruction control. Stochastic formulations,
such as variational multi-scale 3DGS~\cite{li2024variational}, model
distributions over Gaussian parameters and estimate uncertainty from the
resulting samples. Parameter-space methods instead use information measures:
FisherRF~\cite{jiang2023fisherrf} derives uncertainty from Fisher information,
while POp-GS~\cite{wilson2025popgs} uses P-optimality for next-best-view
selection. PRIMU~\cite{gottwald2026primu} predicts novel-view uncertainty from
primitive-based error and coverage features. These methods principally estimate
uncertainty for view selection or uncertainty-map prediction, rather than using
it to control the contribution of primitives during reconstruction.

Most closely related, Galappaththige \emph{et al.}~\cite{galappaththige2026predictive}
propose a post-hoc predictive-photometric uncertainty method. They freeze a
trained 3DGS representation and fit view-dependent per-primitive uncertainty
channels to training-view reconstruction residuals using a regularised
least-squares objective; the learned channels are rasterised into pixel-wise
uncertainty maps. UGOD has a different purpose and learning mechanism. We learn
the uncertainty head jointly with sparse-view reconstruction from Gaussian
attributes and the viewing direction, rather than fitting residuals after
training. Its uncertainty is then used online to modulate opacity before alpha
compositing and to drive detached soft dropout during training. Consequently,
UGOD targets the quality--compactness trade-off of sparse-view reconstruction,
whereas post-hoc predictive uncertainty methods preserve a fixed renderer and
target uncertainty estimation for downstream decisions.

\section{Method}
\subsection{Preliminaries}
\textbf{3D Gaussian Splatting.}
3DGS initialises a set of anisotropic Gaussians from an SfM point cloud. Each
Gaussian is parameterised by its position, rotation, scale, opacity, and
appearance. For a spatial location $\mathbf{x}\in\mathbb{R}^{3}$, a Gaussian
with mean $\boldsymbol{\mu}$ and covariance $\boldsymbol{\Sigma}$ is
\begin{equation}
G(\mathbf{x})=\exp\!\left(-\frac{1}{2}(\mathbf{x}-\boldsymbol{\mu})^\top
\boldsymbol{\Sigma}^{-1}(\mathbf{x}-\boldsymbol{\mu})\right).
\end{equation}
Here $\boldsymbol{\mu}\in\mathbb{R}^{3}$ is the mean and
$\boldsymbol{\Sigma}\in\mathbb{R}^{3\times3}$ is a positive semi-definite
covariance matrix parameterised as
$\boldsymbol{\Sigma}=\mathbf{R}\mathbf{S}\mathbf{S}^{\top}\mathbf{R}^{\top}$,
where $\mathbf{R}$ is an orthogonal rotation matrix and $\mathbf{S}$ is a
diagonal scale matrix. In the implementation, $\mathbf{R}_i$ is parameterised
by a unit quaternion $\mathbf{q}_i$, and $\mathbf{S}_i$ by a scale vector
$\mathbf{s}_i$. We use these compact per-Gaussian parameterisations below.

The Gaussians are projected to the image plane by splatting-based
rasterisation~\cite{zwicker2001surface}. A first-order approximation of the
projection at the Gaussian mean gives the projected covariance
\begin{equation}
    \boldsymbol{\Sigma}^{\prime}=\mathbf{J}\mathbf{W}
    \boldsymbol{\Sigma}\mathbf{W}^{\top}\mathbf{J}^{\top},
\end{equation}
where $\mathbf{J}$ is the Jacobian of the local projective transformation and
$\mathbf{W}$ is the view transformation matrix. For pixel coordinate
$\mathbf{x}$, let $G_i^{2\mathrm{D}}(\mathbf{x})$ denote the projected
footprint of the $i$-th Gaussian and $o_i\in(0,1)$ its learned opacity. Its
screen-space compositing weight and transmittance are
\begin{equation}
\alpha_i(\mathbf{x})=o_iG_i^{2\mathrm{D}}(\mathbf{x}),\qquad
T_i(\mathbf{x})=\prod_{j<i}\bigl(1-\alpha_j(\mathbf{x})\bigr),
\label{eq:alpha_compositing}
\end{equation}
where Gaussians are ordered from front to back. The rendered RGB colour is
\begin{equation}
\mathbf{C}(\mathbf{x})=\sum_{i=1}^{n}
T_i(\mathbf{x})\,\alpha_i(\mathbf{x})\,\mathbf{c}_i.
\label{eq:rendering}
\end{equation}
\begin{figure*}[t]
    \centering
    \includegraphics[width=1\linewidth]{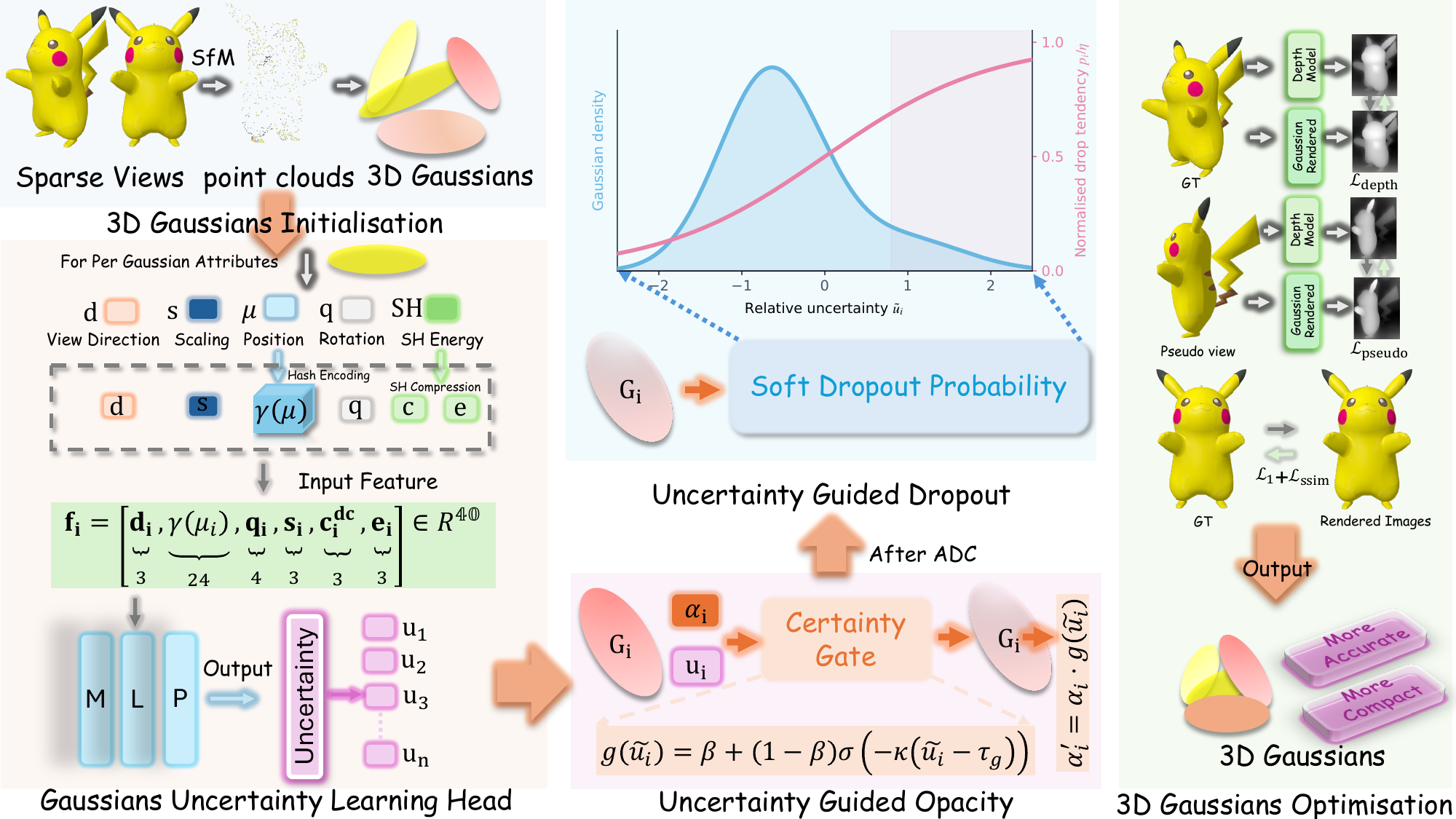}
    \caption{\textbf{Overview of UGOD.} Sparse input views are reconstructed
    by SfM and initialise 3D Gaussian primitives (left). For each visible
    Gaussian, the uncertainty head combines the viewing direction, HashGrid
    position encoding, geometric attributes, and compressed SH appearance to
    predict a view-dependent uncertainty $u_i$ (lower left). This score drives
    two gradient-decoupled mechanisms (centre): normalised uncertainty
    differentiably gates opacity before compositing, whereas its detached copy
    determines the training-only Concrete soft-dropout probability. The
    resulting Gaussians are jointly optimised using supervised depth,
    pseudo-view depth, and photometric losses (right). At inference, soft
    dropout and relative-uncertainty gating are disabled, and deterministic
    raw-uncertainty opacity modulation is used.}
    \label{fig:overview}
\end{figure*}
\textbf{Opacity in 3DGS.}
The learned opacity $o_i$ is optimised jointly with the other Gaussian
attributes. It determines the screen-space weight $\alpha_i$ in
Eq.~\ref{eq:alpha_compositing} and influences density control during training:
primitives with very low opacity are pruned, while the remaining primitives
continue to refine the representation. Under sparse or noisy initialisation,
however, this view-independent scalar cannot distinguish a reliably supported
contribution from one that is ambiguous for the current camera. The following
section introduces view-dependent uncertainty to address this limitation.


\subsection{Per-Gaussian Uncertainty Prediction}
\label{sec:gaussians_uncertainty_learning}

We augment 3D Gaussian Splatting (3DGS)~\cite{kerbl3Dgaussians} with a lightweight
per-Gaussian uncertainty head $\mathcal{M}_\theta$. The predicted uncertainty
serves two coupled but \emph{gradient-decoupled} roles: (i) a differentiable
\emph{opacity-modulation function} that attenuates the opacity of
high-uncertainty Gaussians and
trains the head through the photometric loss, and (ii) a detached, training-only
\emph{soft dropout} that acts as a tail regulariser on the least certain
Gaussians. A PSNR-based early-stopping rule freezes the head once it stops
helping, so the learned uncertainty stays meaningful and never destabilises the
underlying geometry.

\paragraph{Notation and overview.}
For each visible Gaussian $i\in\mathcal{V}$, the uncertainty head predicts a
raw uncertainty $u_i$, which is robustly normalised to $\tilde{u}_i$. We denote
by $\breve{u}_i$ the gradient-detached copy of $\tilde{u}_i$: it equals
$\tilde{u}_i$ in the forward pass but satisfies
$\partial \breve{u}_i / \partial \tilde{u}_i = 0$ during back-propagation.
The normalised value drives a differentiable opacity-modulation factor $g_i$,
while $\breve{u}_i$ drives a training-only dropout probability $p_i$ and soft
keep mask $m_i$. Thus, the uncertainty pathway is
$u_i\rightarrow\tilde{u}_i\rightarrow g_i$ and
$\breve{u}_i\rightarrow p_i\rightarrow m_i$.

\subsubsection{Uncertainty Head: Inputs and Architecture}
\label{sec:uncertainty_head}
Each Gaussian $G_i$ carries a mean $\boldsymbol{\mu}_i\!\in\!\mathbb{R}^{3}$, a
rotation quaternion $\mathbf{q}_i\!\in\!\mathbb{R}^{4}$, an (activated) scale
$\mathbf{s}_i\!\in\!\mathbb{R}^{3}$, a DC colour
$\mathbf{c}^{\mathrm{dc}}_i\!\in\!\mathbb{R}^{3}$, and higher-order spherical
harmonic (SH) coefficients. Concretely, for SH degree $D$ we denote the
(non-DC) coefficients as $\{\mathbf{c}^{(\ell)}_i\}_{\ell=1}^{L}$ with
$\mathbf{c}^{(\ell)}_i\!\in\!\mathbb{R}^{3}$ and $L=D(D+2)$.

\paragraph{SH compression.}
Directly feeding all SH coefficients to the uncertainty head would introduce a
high-dimensional, highly correlated appearance input: the uncertainty prediction
should depend on how ``view-dependent'' a Gaussian is, not on the specific
orientation of its SH lobes. We therefore keep the DC colour
$\mathbf{c}^{\mathrm{dc}}_i$ unchanged and summarise the remaining SH bands by a
per-channel magnitude statistic. 

Specifically, we compute a \emph{rest-energy} vector $\mathbf{e}_i\in\mathbb{R}^3$
by averaging the element-wise absolute values of the non-DC coefficients across
all bands:
\begin{equation}
\mathbf{e}_i 
\;=\; \frac{1}{L}\sum_{\ell=1}^{L}\bigl|\mathbf{c}^{(\ell)}_i\bigr|,
\end{equation}
where $|\cdot|$ is applied element-wise to the RGB channels. Intuitively,
$\mathbf{e}_i$ measures the overall strength of view-dependent effects per colour
channel while discarding directional details. This reduces the SH block from
$3L$ scalars (e.g., $3\times 15=45$ values for degree $D\!=\!3$) to $3$ scalars.
Together with the DC colour, we obtain a compact $6$-D appearance descriptor
$[\mathbf{c}^{\mathrm{dc}}_i,\mathbf{e}_i]$ used as part of the head input.

\paragraph{Input feature.}
Given the camera centre $\mathbf{o}$, we form the unit view direction
$\mathbf{d}_i=(\boldsymbol{\mu}_i-\mathbf{o})/\lVert\boldsymbol{\mu}_i-\mathbf{o}\rVert$
and a multi-resolution hash encoding $\gamma(\boldsymbol{\mu}_i)\!\in\!\mathbb{R}^{24}$
of the position. The head input is the $40$-dimensional concatenation
\begin{equation}
\mathbf{f}_i =
\bigl[\mathbf{d}_i,\,
\gamma(\boldsymbol{\mu}_i),\,
\mathbf{q}_i,\,
\mathbf{s}_i,\,
\mathbf{c}^{\mathrm{dc}}_i,\,
\mathbf{e}_i\bigr]
\in \mathbb{R}^{40}.
\end{equation}
The respective dimensions of these components are $3$, $24$, $4$, $3$, $3$,
and $3$.

\paragraph{Network.}
The position is encoded with a multi-resolution hash grid
($6$ levels, $4$ features/level, base resolution $16$, per-level scale $1.5$,
log$_2$ hashmap size $15$), yielding the $24$-D code $\gamma(\cdot)$. The
$40$-D feature is mapped to a scalar logit by a fully fused MLP
$\mathcal{M}_{\theta}:\mathbb{R}^{40}\rightarrow\mathbb{R}$, where $\theta$
denotes its learnable parameters. The MLP has two hidden layers of $32$
neurons, each followed by a LeakyReLU activation. Let
$\sigma(x)=(1+\exp(-x))^{-1}$ denote the logistic sigmoid. We use a single-head
configuration that emits one view-conditioned uncertainty score per visible
Gaussian,
\begin{equation}
\begin{aligned}
u_i &= \sigma\!\bigl(\mathcal{M}_\theta(\mathbf{f}_i)\bigr)\in(0,1),\\
u_i &\leftarrow \operatorname{clip}\!\left(u_i,\,10^{-3},\,1-10^{-3}\right).
\end{aligned}
\end{equation}
This score is not a calibrated epistemic-uncertainty estimate: it is learned
only through the photometric objective via differentiable opacity modulation,
without an uncertainty target or calibration loss.

We robustly normalise the raw uncertainties over the visible Gaussian set
$\mathcal{V}$:
\begin{equation}
\tilde{u}_i =
\operatorname{clip}\!\left(
\frac{u_i-\operatorname{med}_{j\in\mathcal{V}}(u_j)}
{\operatorname{MAD}_{j\in\mathcal{V}}(u_j)+\delta},
[-c,c]\right),
\label{eq:uncertainty_normalization}
\end{equation}
where $\operatorname{med}$ denotes the median, and the median absolute
deviation is
\begin{equation}
\operatorname{MAD}_{j\in\mathcal{V}}(u_j)
= \operatorname{med}_{j\in\mathcal{V}}\!\left|
u_j-\operatorname{med}_{k\in\mathcal{V}}(u_k)\right|.
\end{equation}
This robust scale estimate limits the
influence of a small number of high-uncertainty Gaussians; $\delta>0$ prevents
division by zero and $c=2$ bounds the relative uncertainty. Both the
opacity-modulation function and the dropout module use $\tilde{u}_i$.

\subsubsection{Uncertainty-Guided Opacity Modulation}
\label{sec:opacity_modulation}
We attenuate opacity rather than directly removing Gaussians, so that
unreliable primitives contribute less to alpha compositing while still
receiving photometric supervision. We map relative uncertainty to a
monotonically decreasing, differentiable gate:
\begin{equation}
g_i=\beta+(1-\beta)\,\sigma\!\bigl(-\kappa(\tilde{u}_i-\tau_g)\bigr),
\label{eq:certainty_gate}
\end{equation}
where $\kappa$ controls the transition sharpness, $\tau_g$ specifies the
relative-uncertainty threshold, and $\beta$ is the minimum gate value. Thus,
low-uncertainty Gaussians have $g_i$ close to $1$ and retain their learned
opacity, whereas high-uncertainty Gaussians approach the floor $\beta$. The
floor prevents modulation alone from fully removing a Gaussian. The gate
modulates the learned opacity explicitly as
\begin{equation}
o_i^{\mathrm{gate}}=o_i\,g_i,
\label{eq:gated_opacity}
\end{equation}
so high uncertainty reduces the primitive's screen-space alpha and hence its
colour and transmittance contribution before compositing. The
relative-uncertainty modulation is enabled only after a warm-up of $T_g$
iterations. Before warm-up, and during evaluation, the implementation uses the
base raw-uncertainty opacity operation
\begin{equation}
o_i^{\mathrm{raw}}=(1-u_i)o_i,\qquad
\alpha_i^{\mathrm{raw}}(\mathbf{x})=o_i^{\mathrm{raw}}G_i^{2\mathrm{D}}(\mathbf{x}).
\label{eq:raw_opacity_modulation}
\end{equation}
The two formulations are intentionally assigned different roles. During
optimisation, the relative gate is a distribution-adaptive regulariser: its
median--MAD normalisation makes the threshold responsive to the \emph{relative}
reliability of currently visible Gaussians, even as densification and pruning
change the scale of predicted uncertainties. The floor $\beta$ additionally
prevents premature removal of primitives and preserves useful gradients.
At inference, however, this normalisation would make a Gaussian's opacity
depend on which other primitives happen to be visible in the same view. We
therefore replace the training-time relative gate with the raw-uncertainty rule
in Eq.~\ref{eq:raw_opacity_modulation}, which gives each Gaussian a
deterministic, per-primitive attenuation independent of population statistics.
Thus, relative gating stabilises optimisation, whereas the raw rule provides a
well-defined deployed renderer for all reported evaluations.
The modulated screen-space weight is
\begin{equation}
\alpha_i^{\mathrm{gate}}(\mathbf{x})=
o_i^{\mathrm{gate}}G_i^{2\mathrm{D}}(\mathbf{x}).
\end{equation}
Crucially, $\tilde{u}_i$ is \emph{not} detached on this branch, so the
photometric gradient flows through $g_i$ back into $\mathcal{M}_\theta$:
this is the sole supervisory signal that teaches the head what ``uncertain''
means. The relative-uncertainty modulation is used during training after
warm-up, evaluation uses Eq.~\ref{eq:raw_opacity_modulation}.
\paragraph{Rasterised uncertainty map.}
To visualise the learned per-Gaussian uncertainty in image space, we
rasterise an uncertainty map~$U$ by projecting the predicted
values~$\{u_i\}$ with the same $\alpha$-blending pipeline used for RGB,
while keeping the Gaussian geometry fixed.
Formally, for a pixel~$\mathbf{x}$,
\begin{equation}
\label{eq:uncertainty_raster}
U(\mathbf{x})
=
\sum_{i}
T_i(\mathbf{x})\,\alpha_i(\mathbf{x})\,u_i,
\end{equation}
where $T_i$ and $\alpha_i$ follow the standard 3DGS compositing
(Eq.~\ref{eq:alpha_compositing}); thus, $U$ uses unmodulated compositing
weights and is distinct from the training and inference weights defined in Supplementary Eq.~(11).
This map is used only for analysis. During training, opacity is modulated by
the relative-uncertainty function after warm-up, whereas soft dropout remains
training-only.
\subsubsection{Detached Soft Dropout (Tail Regulariser)}
\label{sec:soft_dropout}
Opacity modulation and soft dropout have complementary roles. The opacity gate
is a deterministic rendering mechanism: on every training view, it
continuously down-weights an uncertain Gaussian while preserving its
contribution and the gradient that learns its uncertainty. This alone does not
prevent the representation from repeatedly relying on the same weakly
constrained Gaussians. Soft dropout instead acts as a stochastic
training-only regulariser: occasionally suppressing these Gaussians forces the
remaining, better-supported primitives to explain the observations and reduces
co-adaptation to unreliable evidence. Accordingly, it assigns greater
regularisation to high-uncertainty Gaussians. The per-Gaussian drop probability
uses the gradient-detached uncertainty $\breve{u}_i$ defined above,
\begin{equation}
p_i=r(t)\,\eta\,\sigma(\breve{u}_i-\tau_d),
\label{eq:drop_probability}
\end{equation}
where $\eta$ scales the uncertainty influence, $\tau_d$ is a drop threshold,
and $r(t)=\mathrm{clip}\!\big((t-T_d)/R,\,0,\,1\big)$ is a linear ramp that
starts dropout after a fixed warm-up at $T_d=1200$ and reaches its full value
over $R=500$ iterations. Before applying the logit, we clip
$p_i\leftarrow\mathrm{clip}(p_i,\varepsilon_p,1-\varepsilon_p)$ with
$\varepsilon_p>0$. We then convert $p_i$ into a differentiable
\emph{concrete} keep mask. We first form the perturbed logit
\[
z_i=\mathrm{logit}(p_i)+\mathrm{logit}(\varepsilon_i),
\]
where $\varepsilon_i\sim\mathcal{U}(\varepsilon_p,1-\varepsilon_p)$ is
independent uniform noise. The keep mask is then
\begin{equation}
m_i=\mathrm{clip}_{[m_{\min},\,m_{\max}]}\!\left[
1-\sigma\!\left(\frac{z_i}{T_{\mathrm{conc}}}\right)\right],
\label{eq:concrete_mask}
\end{equation}

where $T_{\mathrm{conc}}$ is the temperature. We apply the mask as
$\alpha_i^{\mathrm{train}}(\mathbf{x})=
\alpha_i^{\mathrm{gate}}(\mathbf{x})m_i$. This stochastic masking is useful
only during optimisation: by requiring the scene to remain explainable under
occasional removal of uncertain primitives, it discourages co-adaptation and
over-reliance on weakly constrained evidence. At test time, the objective is
instead to render the best deterministic estimate of the complete scene.
Consequently, we disable dropout and retain every primitive. This does not
discard uncertainty: a high score indicates comparatively unreliable, rather
than certainly invalid, evidence, so hard removal could eliminate useful
detail. Rendering therefore uses
$\alpha_i^{\mathrm{raw}}(\mathbf{x})$ from Eq.~\ref{eq:raw_opacity_modulation}
to deterministically attenuate each uncertain primitive in proportion to its
score, rather than randomly dropping it.

\paragraph{Gradient flow.}
The two mechanisms share one scalar $u_i$ but are deliberately decoupled:
the opacity-modulation branch is differentiable and back-propagates the rendering loss into
$\mathcal{M}_\theta$, whereas the dropout branch consumes $\breve{u}_i$
and therefore \emph{never} reshapes the uncertainty. This prevents the
degenerate solution in which the head inflates $u_i$ merely to be dropped, and
lets the head learn a photometrically useful uncertainty score while dropout remains a
pure regulariser. Because $\breve{u}_i$ is gradient-detached, the
dropout-probability path does not update the uncertainty head; it does not detach the rendered mask from the
Gaussian parameters. Consequently, the Concrete mask regularises Gaussian
opacity, geometry, and appearance through the rendered photometric loss.

\subsubsection{Training Objective}
Following FSGS~\cite{zhu2024fsgs}, Full UGOD adopts DPT-based monocular depth
regularisation~\cite{ranftl2021vision} to stabilise geometry under sparse
observations; it is an adopted training component rather than a contribution of
UGOD. A frozen DPT-Hybrid estimator
predicts a depth prior $D$ once for each training image during dataset loading.
Let $\hat{D}$ be the rasterised depth and $\rho(\cdot,\cdot)$ denote Pearson
correlation. To accommodate the scale and sign ambiguity of monocular depth,
we use
\begin{equation}
\begin{aligned}
\mathcal{L}_{\mathrm{depth}}
={}&\min\!\left\{
1-\rho(-D,\hat{D}),\right.\\[-2pt]
&\left.\hspace{17mm}1-\rho\!\left((D+200)^{-1},\hat{D}\right)
\right\}.
\end{aligned}
\label{eq:depth_loss}
\end{equation}
Between iterations $500$ and $5500$, we additionally sample pseudo cameras,
render RGB and depth, and apply the frozen DPT-Hybrid estimator to the rendered
RGB. This provides a pseudo-view depth prior for regularising geometry beyond
the observed training views. The pseudo-view depth loss is
\begin{equation}
\mathcal{L}_{\mathrm{pseudo}}=
1-\rho\!\left(\hat{D}_{\mathrm{p}},-D_{\mathrm{p}}\right),
\label{eq:pseudo_depth_loss}
\end{equation}
where $D_{\mathrm{p}}$ is the DPT prediction for the rendered pseudo view and
$\hat{D}_{\mathrm{p}}$ is its rasterised depth. This term is linearly ramped
during its first $500$ active iterations. DPT is a fixed target: gradients
from $\mathcal{L}_{\mathrm{pseudo}}$ flow through the rendered depth, not
through the DPT preprocessing path.

The Gaussians and uncertainty head are jointly optimised with
\begin{equation}
\begin{aligned}
\mathcal{L}={}&(1-\lambda)\,\mathcal{L}_1+
\lambda\,\bigl(1-\mathrm{SSIM}\bigr)\\
&+\beta_d(t)\mathcal{L}_{\mathrm{depth}}
+\alpha(t)\lambda_p\mathcal{L}_{\mathrm{pseudo}},
\end{aligned}
\label{eq:photometric_loss}
\end{equation}
where $\lambda=0.2$, $\beta_d(t)=0.05$ through iteration $5500$ and
$0.001$ thereafter, and $\alpha(t)$ is the pseudo-view ramp. We use
$\lambda_p=0.5$ by default and $\lambda_p=0.03$ for outdoor MiP-NeRF~360
scenes.
The uncertainty head is supervised implicitly through the differentiable
opacity-modulation function in Eq.~\ref{eq:certainty_gate}; we use no auxiliary
uncertainty-calibration loss in the reported experiments. The head uses its own
Adam optimiser (cosine-annealed learning rate), independent of the 3DGS
optimiser. We apply validation-based early stopping to the uncertainty head;
its criterion and hyper-parameters are provided in the supplementary material.

The complete training and inference procedure is provided in
Supplementary Algorithm~1.

\paragraph{Default hyper-parameters.}
The complete uncertainty-head, gating, dropout, and early-stopping settings
are reported in Supplementary Table~1.

\section{Experiments}
\label{Experiments}
\subsection{Experimental Setup}
\textbf{Datasets and protocols.} We evaluate sparse-view NVS on
MiP-NeRF~360~\cite{barron2022mip} and LLFF~\cite{mildenhall2019llff}. We
follow FSGS~\cite{zhu2024fsgs} for its 24-view MiP-NeRF~360 and 3-view LLFF
protocols, and additionally evaluate the more challenging 8-view MiP-NeRF~360
setting. For MiP-NeRF~360, we use the six scenes reported in
Supplementary Tables~2 and~3, using
predefined disjoint training and evaluation cameras.

For LLFF, we use three COLMAP-reconstructed training views per scene and
evaluate on the corresponding held-out views, as detailed in
Supplementary Table~4. LLFF images are downsampled by a factor of eight
along both spatial dimensions. For MiP-NeRF~360, images wider than 1600 pixels
are resized to width 1600 while preserving aspect ratio, whereas smaller images
retain their original resolution. These camera splits and image resolutions
define our evaluation protocol; consequently, absolute metrics need not match
prior reports that use different splits or input resolutions.

\textbf{Implementation.} We implement UGOD in PyTorch on an NVIDIA RTX A100
and train all benchmarked methods for 6,000 iterations. The uncertainty head
uses the $(6,0,0,0)$ HashGrid configuration selected by the ablation in
Table~\ref{tab:hashgrid_ablation}; its remaining settings follow
Supplementary Table~1. We reset opacity at iterations $2001$ and $5001$ for $6000$-iteration
training. The uncertainty head is frozen after two consecutive validation
evaluations for which PSNR decreases relative to the preceding evaluation
($\Delta\mathrm{PSNR}<\epsilon=0$); Gaussian parameters continue to be
optimised. In tables and comparison figures only, 3DGS* denotes the official
3DGS baseline evaluated under the same sparse-view protocol. Each table
specifies the compared variants:
\emph{Ours (full)} denotes the depth-assisted model with uncertainty-guided
soft dropout.

\textbf{Metrics.} We report PSNR, SSIM~\cite{wang2004image}, and
LPIPS~\cite{zhang2018unreasonable} on held-out views. We additionally report
the final Gaussian count and its reduction relative to 3DGS* and DropGS where
available, since sparse-view reconstruction requires a trade-off between
rendering quality and representation compactness.
\subsection{Quantitative Comparison}
\label{Comparison with Sparse-View Scenes}

\begin{table}[t]
    \centering
    \caption{Scene-average results on LLFF (3 views) and MiP-NeRF~360
    (8/24 views). Reduction is relative to 3DGS*; best and second-best
    values are bold and underlined.}
    \label{tab:average_results}
    \setlength{\tabcolsep}{2.2pt}
    \renewcommand{\arraystretch}{1.08}
    \scriptsize
    \begin{tabular}{@{}llccccc@{}}
        \toprule
        Setting & Method & PSNR$\uparrow$ & SSIM$\uparrow$ & LPIPS$\downarrow$ & \#Gaussians & vs 3DGS* \\
        \midrule
        \multirow{4}{*}{\shortstack[l]{LLFF\\(3-view)}}
        & 3DGS*              & 14.6543 & 0.4379 & 0.3990 & 254{,}350 & -- \\
        & DropGS             & 14.6443 & 0.4557 & 0.3894 & 210{,}414 & 1.2$\times$ \\
        & FSGS (w/ depth)    & \second{19.0457} & \second{0.6230} & \second{0.2514} & \second{151{,}173} & \second{1.7$\times$} \\
        \rowcolor{oursrow}
        & Ours (full)        & \best{19.3300} & \best{0.6370} & \best{0.2449} & \best{102{,}280} & \best{2.5$\times$} \\
        \midrule
        \multirow{4}{*}{\shortstack[l]{Mip360\\(8-view)}}
        & 3DGS*              & 11.9591 & 0.2747 & \second{0.6095} & 864{,}393 & -- \\
        & DropGS             & 12.7267 & 0.3193 & \best{0.6026} & 716{,}133 & 1.2$\times$ \\
        & FSGS (w/ depth)    & \second{13.0309} & \second{0.3588} & 0.6197 & \second{132{,}090} & \second{6.5$\times$} \\
        \rowcolor{oursrow}
        & Ours (full)        & \best{14.1435} & \best{0.3682} & 0.6179 & \best{95{,}662} & \best{9.0$\times$} \\
        \midrule
        \multirow{4}{*}{\shortstack[l]{Mip360\\(24-view)}}
        & 3DGS*              & 18.6739 & \second{0.5614} & \best{0.4289} & 1{,}038{,}952 & -- \\
        & DropGS             & \second{18.9911} & \best{0.5629} & \second{0.4520} & 775{,}653 & 1.3$\times$ \\
        & FSGS (w/ depth)    & 18.7833 & 0.5315 & 0.5245 & \second{148{,}296} & \second{7.0$\times$} \\
        \rowcolor{oursrow}
        & Ours (full)        & \best{19.0867} & 0.5365 & 0.5203 & \best{129{,}498} & \best{8.0$\times$} \\
        \bottomrule
    \end{tabular}
\end{table}
We quantitatively compare Full UGOD with 3DGS*~\cite{kerbl20233d},
DropGS~\cite{park2025dropgaussian}, and FSGS~\cite{zhu2024fsgs}.
Table~\ref{tab:average_results} first reports scene-averaged results, and
Supplementary Tables~2, 3, and~4 provide the corresponding per-scene comparisons.

\textbf{LLFF.} On the more challenging 3-view protocol, Full UGOD achieves
the best scene-average PSNR (19.33), SSIM (0.637), and LPIPS (0.245), while
using 102,280 Gaussians on average $2.5\times$ fewer than 3DGS*. It obtains
the highest PSNR on six of seven scenes and the smallest representation on six
scenes. \emph{Orchids} is the exception for rendering quality, where FSGS
achieves the best PSNR and LPIPS (with tied best SSIM), while on \emph{Room}
FSGS uses fewer Gaussians and achieves the best SSIM and LPIPS. These
per-scene results show that the gain remains scene dependent, but is most
consistent under severe view sparsity.

\textbf{MiP-NeRF 360.} Under the 8-view protocol, Full UGOD obtains the
highest average PSNR (14.14) and SSIM (0.368), with 95,662 final Gaussians on
average. This is $9.0\times$ fewer Gaussians than 3DGS* and $7.5\times$ fewer
than DropGS. The per-scene results show the highest PSNR on five of the six
scenes and the smallest representation on three; on \emph{bicycle}, for
example, Full UGOD improves PSNR from 11.01 to 13.11 while reducing the
Gaussian count from 1,460,812 to 59,483. DropGS achieves the best average
LPIPS, and FSGS attains the best PSNR, SSIM, and LPIPS on \emph{counter};
therefore, the 8-view result is not a uniform improvement for every metric.

With 24 views, Full UGOD still gives the highest average PSNR (19.09) and the
smallest average representation (129,498 Gaussians, $8.0\times$ fewer than
3DGS*). It has the lowest Gaussian count on five of the six scenes and the
highest PSNR on \emph{bicycle}, \emph{counter}, and \emph{kitchen}. The detailed
per-scene analysis in Supplementary Table~3 indicates
that, with more observations, uncertainty attenuation can trade fine detail for
representation compactness. Thus, in this less-sparse setting, the results
demonstrate a quality--compactness trade-off rather than across-the-board metric
superiority.

\subsection{Ablation Study}
\begin{table}[t]
    \centering
    \caption{Scene-average component ablation. $N$ is the number of scenes
    averaged; the 8-view soft-dropout variants use five scenes, while the
    full model uses all six. Best and second-best values are bold and
    underlined.}
    \label{tab:ablation_average}
    \setlength{\tabcolsep}{2.5pt}
    \renewcommand{\arraystretch}{1.08}
    \scriptsize
    \begin{tabular}{@{}llrcccc@{}}
        \toprule
        Setting & Variant & $N$ & PSNR$\uparrow$ & SSIM$\uparrow$ & LPIPS$\downarrow$ & \#Gaussians$\downarrow$ \\
        \midrule
        \multirow{3}{*}{\shortstack[l]{LLFF\\(3-view)}}
        & Ours & 7 & \second{18.2386} & \second{0.5989} & \second{0.2701} & \second{180,905} \\
        & w/o soft dropout & 7 & 17.9529 & 0.5700 & 0.2877 & 183,460 \\
        \rowcolor{oursrow}
        & Ours (full) & 7 & \best{19.3300} & \best{0.6370} & \best{0.2449} & \best{102,280} \\
        \midrule
        \multirow{3}{*}{\shortstack[l]{Mip360\\(8-view)}}
        & w/o soft dropout & 5 & \second{13.9438} & \second{0.3672} & \best{0.6074} & 108,008 \\
        & w/ soft dropout & 5 & 12.9741 & 0.3475 & 0.6267 & \second{102,644} \\
        \rowcolor{oursrow}
        & Ours (full) & 6 & \best{14.1435} & \best{0.3682} & \second{0.6179} & \best{95,662} \\
        \midrule
        \multirow{3}{*}{\shortstack[l]{Mip360\\(24-view)}}
        & w/o soft dropout & 6 & \second{19.0752} & \best{0.5521} & \best{0.4744} & 243,836 \\
        & w/ soft dropout & 6 & 18.7457 & \second{0.5429} & \second{0.4760} & \second{244,322} \\
        \rowcolor{oursrow}
        & Ours (full) & 6 & \best{19.0867} & 0.5365 & 0.5203 & \best{129,498} \\
        \bottomrule
    \end{tabular}
\end{table}

\textbf{Component ablation.}
Table~\ref{tab:ablation_average} summarises the
component rows in Supplementary Tables~5, 6, and~7: for each sparse-view
protocol, it compares uncertainty-guided opacity modulation with and without
soft dropout, and the full configuration, in terms of scene-average PSNR,
SSIM, LPIPS, and final Gaussian count. In the 8-view
setting, the soft-dropout comparison is averaged over the five scenes where
both variants are available; the bicycle scene instead reports modulation-only and
full results. Adding soft dropout alone reduces the average Gaussian count
from 108,008 to 102,644 but also lowers PSNR from 13.94 to 12.97. The full
model obtains the best 8-view average PSNR and SSIM while using 95,662
Gaussians; because it also includes depth supervision, this comparison measures
the complete system rather than an isolated dropout effect. At 24 views, the
full model improves PSNR and compactness but not average SSIM or LPIPS. In
contrast, under the LLFF 3-view protocol, the full configuration improves all
three image metrics and reduces the average Gaussian count to 102,280. These
results support using the complete configuration in the most severely sparse
setting while making the quality--compactness trade-off explicit elsewhere.

\textbf{HashGrid input configuration.} We conduct a controlled ablation on
the MiP-NeRF 360 \emph{kitchen} scene with 24 training views, holding all
non-encoding settings fixed (Table~\ref{tab:hashgrid_ablation}). $P$, $S$,
$R$, and $V$ denote the number of HashGrid levels assigned to position, scale,
rotation, and view direction, respectively. Position-only encoding with six
levels, namely $(P,S,R,V)=(6,0,0,0)$, achieves the best PSNR (15.0757), SSIM
(0.4628), and LPIPS (0.6396). Adding one view-direction level to the
five-level position encoding improves the result over $(5,0,0,0)$, but remains
inferior to $(6,0,0,0)$. Allocating levels to scale and rotation, or increasing
the position encoding to seven levels, degrades all three metrics in this
controlled setting. We therefore use $(P,S,R,V)=(6,0,0,0)$: only position is
HashGrid encoded, while the remaining attributes are concatenated in their raw
form.

\begin{figure*}[t]
  \centering
  \includegraphics[width=\textwidth]{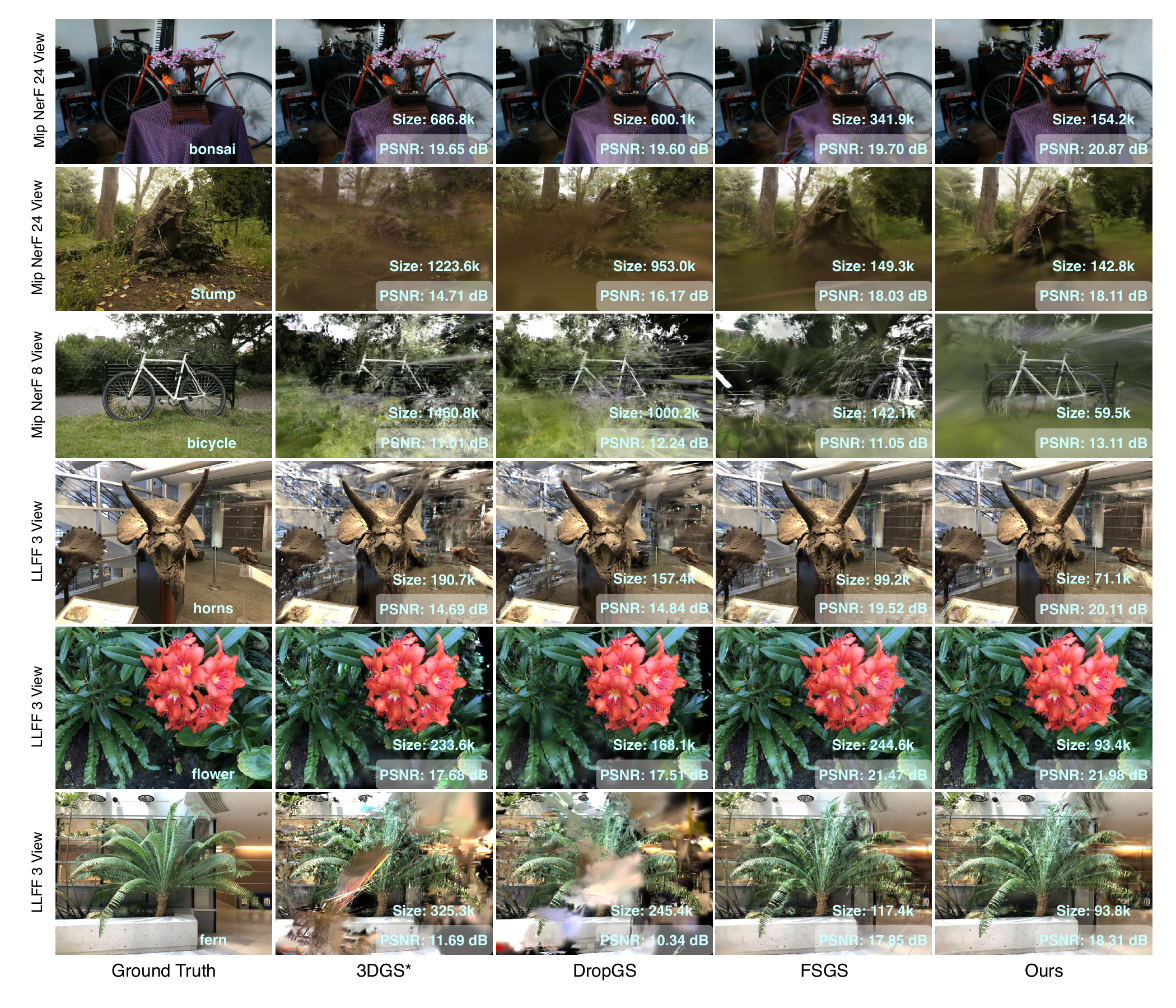}
  \caption{Qualitative comparison on MiP-NeRF~360 (24/8 views) and LLFF
  (3 views). Each non-GT panel reports the displayed-view PSNR and final
  Gaussian count.}
  \label{fig:qualitative}
\end{figure*}

\begin{table}[t]
\centering
\caption{Controlled HashGrid input-encoding ablation on MiP-NeRF 360
\emph{kitchen} with 24 training views. $P$, $S$, $R$, and $V$ are the numbers
of encoding levels assigned to position, scale, rotation, and view direction. \textbf{Bold} indicates the best result.}
\label{tab:hashgrid_ablation}
\setlength{\tabcolsep}{4pt}
\renewcommand{\arraystretch}{1.08}
\begin{tabular}{@{}ccccccc@{}}
\toprule
$P$ & $S$ & $R$ & $V$ & PSNR $\uparrow$ & SSIM $\uparrow$ & LPIPS $\downarrow$ \\
\midrule
5 & 0 & 0 & 0 & 12.8370 & 0.4092 & 0.6619 \\
5 & 0 & 0 & 1 & 14.7173 & 0.4575 & 0.6400 \\
\rowcolor{oursrow}
6 & 0 & 0 & 0 & \textbf{15.0757} & \textbf{0.4628} & \textbf{0.6396} \\
6 & 1 & 1 & 0 & 14.6207 & 0.4508 & 0.6461 \\
7 & 0 & 0 & 0 & 10.9079 & 0.4070 & 0.6758 \\
\bottomrule
\end{tabular}
\end{table}

\subsection{Qualitative Comparison}
\label{sec:qualitative}

Figure~\ref{fig:qualitative} presents qualitative comparisons on
Mip-NeRF~360 (24-view \emph{stump} and 8-view \emph{bicycle}) and
LLFF (3-view \emph{horns}, \emph{flower}, and \emph{fern}).
Each rendered view is annotated with the displayed-view PSNR and the final
Gaussian count of the corresponding model.

On the 8-view \emph{bicycle} scene, 3DGS* and DropGS exhibit substantial
floaters and blurred structures, whereas FSGS recovers more coherent geometry.
Full UGOD more clearly recovers bicycle spokes and the background fence in this
example, with a view-wise PSNR of 13.11\,dB and 59.5K Gaussians, compared with
11.01\,dB and 1460.8K for 3DGS*, 12.24\,dB and 1000.2K for DropGS, and
11.05\,dB and 142.1K for FSGS.

The same quality--compactness pattern is visible in the LLFF examples. On
\emph{horns}, \emph{flower}, and \emph{fern}, Full UGOD attains higher
displayed-view PSNR than FSGS (20.11 vs.\ 19.52\,dB, 21.98 vs.\ 21.47\,dB, and
18.31 vs.\ 17.85\,dB, respectively) while using fewer Gaussians (71.1K vs.\
99.2K, 93.4K vs.\ 244.6K, and 93.8K vs.\ 117.4K, respectively). The
24-view \emph{stump} example shows a smaller visual and numerical margin:
Full UGOD reaches 18.11\,dB with 142.8K Gaussians, compared with
18.03\,dB and 149.3K for FSGS. These examples are consistent with the
quantitative results: the proposed model can improve the quality--compactness
trade-off under sparse-view supervision, although the margin varies across
scenes.

\subsection{Uncertainty Dynamics and Opacity Response}
\label{sec:uncertainty_dynamics}

\begin{figure*}[t]
  \centering
  \includegraphics[width=\textwidth]{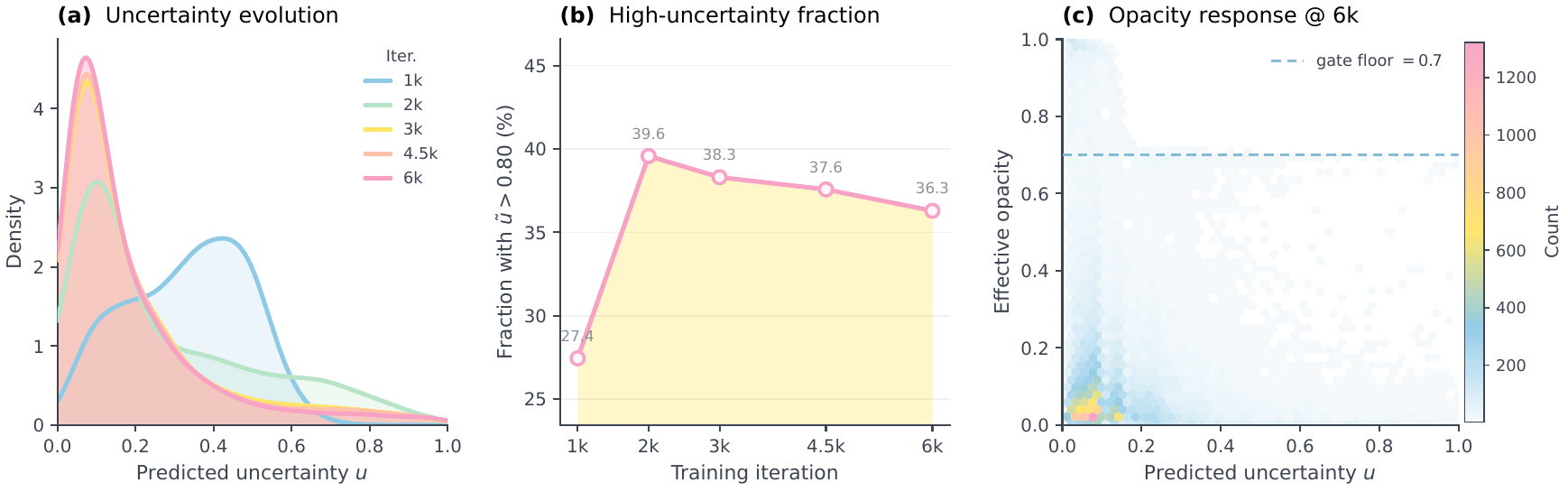}
  \caption{Uncertainty dynamics for the LLFF \emph{fern} scene with three
  training views:
  raw-uncertainty distributions (left), the fraction with
  $\tilde{u}_i>0.8$ (centre), and the uncertainty--opacity response at
  6k iterations (right).}
  \label{fig:uncertainty_dynamics}
\end{figure*}

To characterise the signal learned through the photometric objective, we track
the uncertainty head throughout a representative three-view training run on
the \emph{fern} scene from LLFF (Fig.~\ref{fig:uncertainty_dynamics}). This
analysis concerns the evolution and use of the predicted uncertainty, rather
than calibration error. The raw uncertainty distribution initially has a broad
mode near $u=0.45$, but
progressively concentrates near $u=0.1$ while retaining a high-uncertainty
tail (Fig.~\ref{fig:uncertainty_dynamics}a). Thus, training does not uniformly
reduce all uncertainty predictions: most Gaussians become low-uncertainty,
whereas a subset remains uncertain.

The proportion of relative uncertainties satisfying $\tilde{u}_i>0.8$ rises
from approximately $27\%$ at 1k iterations to $40\%$ at 2k, coinciding with
the densification period, and then decreases to approximately $36\%$ at 6k
(Fig.~\ref{fig:uncertainty_dynamics}b). This trajectory is consistent with
densification introducing initially less-constrained Gaussians, after which
photometric optimisation separates the low-uncertainty majority from a
persistent uncertain tail. At 6k iterations, larger predicted uncertainties
are associated with smaller gate multipliers (Fig.~\ref{fig:uncertainty_dynamics}c).
The response approaches the configured multiplier floor $\beta=0.70$ for
high-uncertainty Gaussians, directly illustrating their attenuation through
the differentiable opacity-modulation pathway in
Eq.~\ref{eq:certainty_gate}.

\subsection{Photometric Residuals and Uncertainty Visualisation}
\label{sec:uncertainty_vis}
\begin{figure*}[!t]
  \centering
  \includegraphics[width=\textwidth]{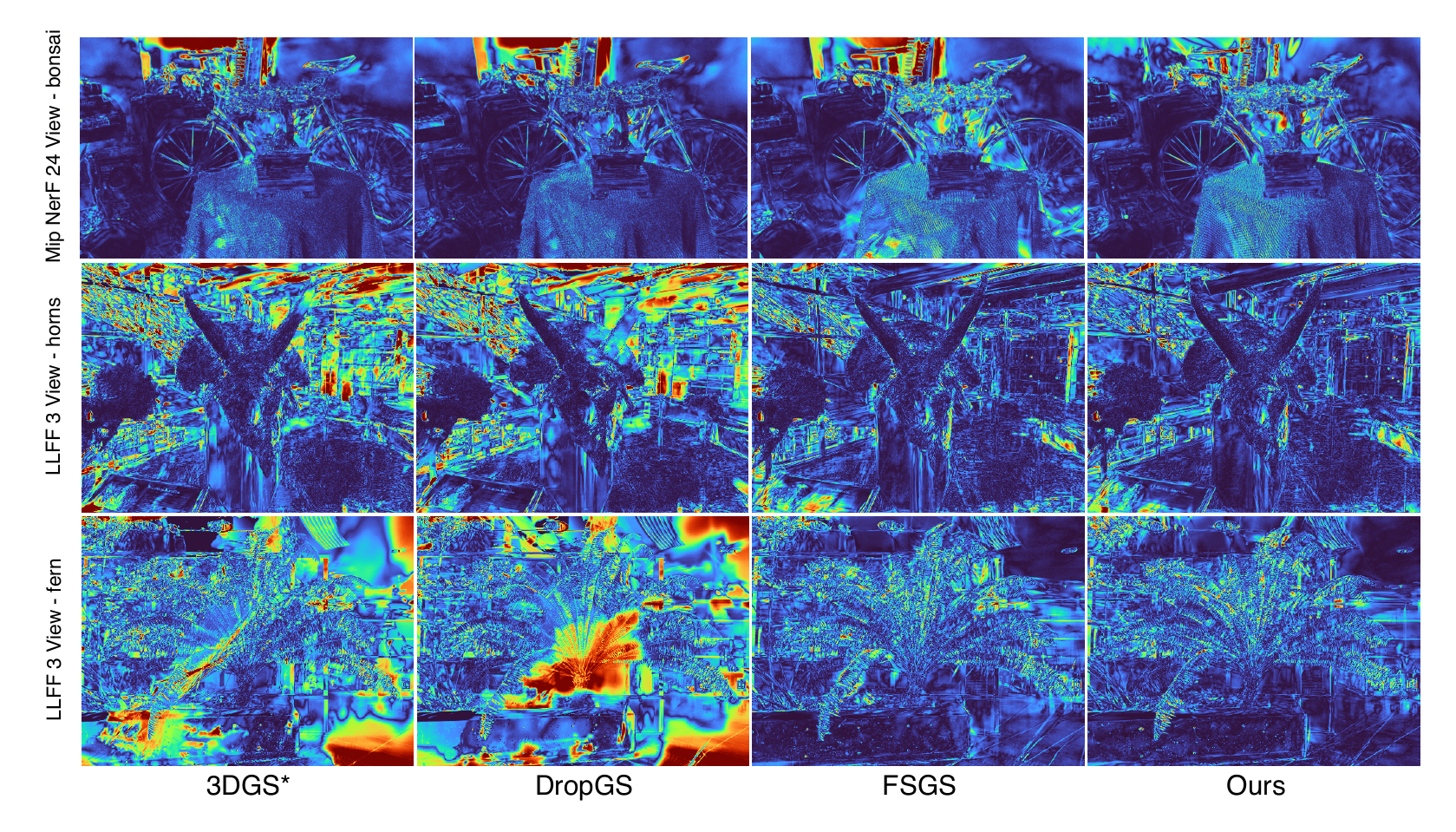}
    \caption{Photometric residuals for 3DGS*, DropGS, and FSGS, and UGOD's
  predicted uncertainty. Rows show MiP-NeRF~360 \emph{bonsai} (24 views)
  and LLFF \emph{horns} and \emph{fern} (3 views).}
  \label{fig:uncertainty_maps}
\end{figure*}
Only UGOD contains the uncertainty head and can therefore produce the
predicted uncertainty map $U(\mathbf{x})$ in
Eq.~\ref{eq:uncertainty_raster}. Its scalar $u_i$ is a view-dependent signal
predicted from Gaussian attributes and viewing direction, and is only
indirectly supervised through the opacity-modulation pathway.

For 3DGS*, DropGS, and FSGS, Figure~\ref{fig:uncertainty_maps} instead shows
the photometric residual map
\begin{equation}
E(\mathbf{x})=
\left\lVert\hat{\mathbf{C}}(\mathbf{x})-\mathbf{C}_{\mathrm{GT}}(\mathbf{x})\right\rVert_1.
\label{eq:photometric_residual}
\end{equation}
These baseline heatmaps measure image-space reconstruction error, not
predicted uncertainty. The visualisation consequently compares residual
patterns for the baselines with the predicted uncertainty pattern of UGOD on
MiP-NeRF~360 \emph{bonsai} (24 views) and LLFF \emph{horns} and
\emph{fern} (three training views).
The baseline residuals are concentrated around under-constrained structures
and backgrounds, whereas UGOD's predicted map assigns lower accumulated
uncertainty to most retained scene content.

\section{Conclusion}
We introduced UGOD, an uncertainty-guided framework for sparse-view 3D
Gaussian Splatting. UGOD learns view-dependent per-Gaussian uncertainty from
Gaussian attributes and the viewing direction, then uses this signal in two
complementary mechanisms. First, uncertainty-guided opacity modulation
continuously attenuates high-uncertainty contributions before alpha
compositing. Second, gradient-detached soft dropout regularises
high-uncertainty primitives during training without directly altering the
uncertainty estimator. Experiments on MiP-NeRF 360 and LLFF show that this
design improves the quality--compactness trade-off in the evaluated sparse-view
protocols, with the most consistent gains under severe view sparsity. The
per-scene results further show that the benefit is metric and scene dependent,
especially when additional input views reduce the reconstruction ambiguity.

\backmatter

\bmhead{Acknowledgements}
This work was supported by Manchester Metropolitan University Student Scholarship.

\bibliography{ugod}

\clearpage
\setcounter{section}{0}
\setcounter{subsection}{0}
\setcounter{subsubsection}{0}
\setcounter{table}{0}
\setcounter{figure}{0}
\setcounter{equation}{0}
\setcounter{algorithm}{0}
\renewcommand{\thesection}{S\arabic{section}}
\begin{center}
{\LARGE\bfseries Supplementary Information}\\[0.5em]
{\large UGOD: Uncertainty-Guided Opacity and Dropout for Sparse-View 3D Gaussian Splatting}
\end{center}
\vspace{1em}

\section{Supplementary Material}\label{sec:appendix}

This appendix supplements the main paper with four blocks of material:
supplementary mathematical derivations, implementation details, per-scene
quantitative comparisons, and component-ablation tables for the uncertainty-guided
operators in Sec.~3.2 of the main paper.
We present the mathematical analysis first, followed by additional experimental
results and implementation details.

\subsection{Supplementary Mathematical Analysis}
\label{sec:theoretical_motivation}

This subsection provides supplementary derivations for the robust uncertainty
normalisation, alpha-compositing sensitivity, opacity modulation, detached
Concrete dropout, depth-regularisation schedule, and train--test opacity
weighting.
These details explain the behaviour of the operators used in the main method
without introducing an additional training objective.

\subsubsection{SH Compression: Compact Appearance Descriptor}
The rest-energy vector $\mathbf{e}_i$ is defined in the main text as the
channel-wise $L_1$ mean of the non-DC SH coefficients. It is therefore a compact
magnitude surrogate for view-dependent appearance, rather than an exact
spectral-energy or Parseval-equivalent quantity. This summary retains the
overall strength of the non-DC coefficients while avoiding a high-dimensional,
orientation-specific appearance input to the uncertainty head.

\subsubsection{The Input Feature Vector: Manifold Coordinate Embedding}

The feature vector $\mathbf{f}_i$ is defined in the main text by concatenating
view direction, hashed position, rotation, scale, DC colour, and
rest-energy. The multi-resolution hash encoding $\gamma(\boldsymbol{\mu}_i)$
provides a localised spatial descriptor, while the remaining attributes provide
the Gaussian's view- and appearance-dependent state. Thus, the head learns
uncertainty from both local position and per-Gaussian attributes rather than
from position alone.

\subsubsection{Robust Normalisation: Non-Parametric Order Statistics}

Equation~(8) of the main paper defines the median/MAD
normalisation used in the main method. Denote its median and MAD terms by
$\mu_u$ and $\operatorname{MAD}_{\mathcal{V}}(u)$, respectively. For an
affine reparameterisation $u_i'=a u_i+b$ with $a>0$ and negligible $\delta$,
these terms transform as
\begin{equation}
\begin{aligned}
\mu_{u'} &= a\mu_u+b,\\
\operatorname{MAD}(u') &= a\operatorname{MAD}(u),\\
\bar{u}_i' &= \bar{u}_i.
\end{aligned}
\end{equation}
The modulation function and dropout therefore depend on relative
uncertainty within the visible set rather than the arbitrary absolute scale of
the head output. The positive $\delta$ retains numerical stability when the
visible uncertainties are nearly identical, while clipping limits the
influence of extreme normalised values.

\subsubsection{Alpha-Compositing Sensitivity}
For a fixed depth order and pixel $\mathbf{x}$, denote the colour accumulated
by the Gaussians behind $i$, conditional on unit incoming transmittance, by
\begin{equation}
\mathbf{C}_{>i}(\mathbf{x})=
\sum_{k>i}\left[\prod_{i<j<k}\bigl(1-\alpha_j(\mathbf{x})\bigr)\right]
\alpha_k(\mathbf{x})\mathbf{c}_k .
\label{eq:behind_colour}
\end{equation}
Equation~(4) of the main paper can then be written as
\begin{equation}
\mathbf{C}(\mathbf{x})=T_i(\mathbf{x})\!
\left[\alpha_i(\mathbf{x})\mathbf{c}_i+
\bigl(1-\alpha_i(\mathbf{x})\bigr)\mathbf{C}_{>i}(\mathbf{x})\right].
\label{eq:rendering_decomposition}
\end{equation}
Differentiating with respect to $\alpha_i$ gives
\begin{equation}
\frac{\partial\mathbf{C}(\mathbf{x})}{\partial\alpha_i(\mathbf{x})}
=T_i(\mathbf{x})\bigl(\mathbf{c}_i-\mathbf{C}_{>i}(\mathbf{x})\bigr).
\label{eq:alpha_rendering_gradient}
\end{equation}
Thus, an opacity change affects both the primitive's direct colour
contribution and the transmittance available to all later Gaussians. This
coupling explains why regulating an unreliable primitive before compositing
can affect the rendered pixel even when its own colour is not dominant.

\subsubsection{Opacity-Modulation Bounds and Gradient}

The opacity-modulation function in Eq.~(10) of the main paper maps normalised uncertainty
to a positive opacity multiplier. Because the sigmoid lies in $(0,1)$ and
$0<\beta<1$, the modulation factor is bounded by $\beta<g_i<1$ (or
$\beta\leq g_i\leq1$ in the limiting sense), and is monotone decreasing:
\begin{equation}
\frac{\partial g_i}{\partial\tilde{u}_i}
=-(1-\beta)\kappa\,
\sigma\!\bigl(-\kappa(\tilde{u}_i-\tau_g)\bigr)
\left[1-\sigma\!\bigl(-\kappa(\tilde{u}_i-\tau_g)\bigr)\right]<0 .
\label{eq:gate_derivative}
\end{equation}
For a fixed depth order and projected footprint, the direct gradient path is
\begin{equation}
\frac{\partial\mathcal{L}}{\partial\tilde{u}_i}
=\frac{\partial\mathcal{L}}{\partial o_i^{\mathrm{gate}}}
\frac{\partial o_i^{\mathrm{gate}}}{\partial g_i}
\frac{\partial g_i}{\partial\tilde{u}_i}
=\frac{\partial\mathcal{L}}{\partial o_i^{\mathrm{gate}}}
\,o_i\,\frac{\partial g_i}{\partial\tilde{u}_i}.
\label{eq:gate_gradient}
\end{equation}
The full rendering gradient additionally includes coupling through the
transmittance factors in Eq.~(4) of the main paper. Because the modulation
derivative is negative, a gradient that favours reducing the modulated opacity
increases the corresponding uncertainty. The floor $\beta$ keeps the modulation
factor positive so this
direct gradient path is not removed entirely.

\subsubsection{Detached Soft Dropout}
The dropout probability $p_i$ and the soft keep mask $m_i$ are given in
Eqs.~(15) and~(16) of the main paper. The temperature
$T_{\mathrm{conc}}$ controls the sharpness of the continuous keep-mask
relaxation. Crucially, dropout consumes
$\breve{u}_i$, so its stochastic regularisation does not
directly update the uncertainty head. This prevents the head from increasing
uncertainty merely to obtain a higher dropout probability.
Since $r(t)\in[0,1]$ and the sigmoid lies in $(0,1)$, the probability
satisfies
\begin{equation}
0\leq p_i\leq \eta\,r(t).
\label{eq:drop_probability_bound}
\end{equation}
Writing $z_i=\operatorname{logit}(p_i)+
\operatorname{logit}(\varepsilon_i)$, the unclamped mask is
$m_i=1-\sigma(z_i/T_{\mathrm{conc}})\in(0,1)$. As
$T_{\mathrm{conc}}\rightarrow0$, this continuous relaxation approaches a
binary keep decision; larger temperatures yield smoother gradients. Because
$p_i$ is computed from $\breve{u}_i$,
\begin{equation}
\frac{\partial p_i}{\partial\tilde{u}_i}=0,\qquad
\frac{\partial m_i}{\partial\tilde{u}_i}=0
\quad\text{on the dropout branch}.
\label{eq:dropout_detachment}
\end{equation}
The stochastic branch consequently cannot directly update the uncertainty
head or incentivise it to inflate uncertainty merely to increase the drop
probability.

\subsubsection{Depth-Regularisation Schedule}
For completeness, the Pearson correlation used in
Eqs.~(17) and~(18) of the main paper is computed over
corresponding valid pixels $\Omega$:
\begin{equation}
\rho(\mathbf{a},\mathbf{b})=
\frac{\sum_{q\in\Omega}(a_q-\bar{a})(b_q-\bar{b})}
{\sqrt{\sum_{q\in\Omega}(a_q-\bar{a})^2}\,
\sqrt{\sum_{q\in\Omega}(b_q-\bar{b})^2}},
\label{eq:pearson_correlation}
\end{equation}
where $\bar{a}$ and $\bar{b}$ are the respective pixel-wise means. The
pseudo-view coefficient in Eq.~(19) of the main paper is
\begin{equation}
\alpha(t)=
\begin{cases}
0, & t<500\ \text{or}\ t>5500,\\
(t-500)/500, & 500\leq t<1000,\\
1, & 1000\leq t\leq5500.
\end{cases}
\label{eq:pseudo_ramp}
\end{equation}
Thus, pseudo-view depth regularisation is inactive outside its prescribed
interval and reaches its full weight after a $500$-iteration linear ramp.
Pseudo cameras are drawn from a pre-generated candidate pool produced by the
FSGS random-pose generators for LLFF and $360^\circ$ captures, respectively.
Within the active window they are not applied every iteration: one camera is
sampled every $S$ iterations according to
\texttt{sample\_pseudo\_interval}, after which DPT-Hybrid is evaluated online
on the rendered RGB to form $D_{\mathrm{p}}$.

\subsubsection{Training and Inference Weights}
Combining the warm-up schedule with Eqs.~(10) and~(16) of the main paper, with dropout starting at $T_d=1200$ and ramping for
$R=500$ iterations, the rendering weight is
\begin{equation}
\begin{aligned}
\alpha_i^{\mathrm{train}}(\mathbf{x})
&=
\begin{cases}
\alpha_i^{\mathrm{raw}}(\mathbf{x}), & t<T_g,\\
\alpha_i^{\mathrm{gate}}(\mathbf{x}), & T_g\leq t<T_d,\\
\alpha_i^{\mathrm{gate}}(\mathbf{x})m_i, & t\geq T_d,
\end{cases}\\
\alpha_i^{\mathrm{test}}(\mathbf{x})
&=\alpha_i^{\mathrm{raw}}(\mathbf{x}).
\end{aligned}
\label{eq:train_test_weights}
\end{equation}
This expression makes the train--test distinction explicit: inference uses the
raw-uncertainty opacity operation and disables both relative-uncertainty
modulation and stochastic dropout.

\subsubsection{Validation-Based Early Stopping}
\begin{equation}
\begin{aligned}
\Delta_k&=\mathrm{PSNR}_k-\mathrm{PSNR}^{\mathrm{prev}},\\
c_k&=
\begin{cases}
c_{k-1}+1, & \Delta_k<\epsilon,\\
0, & \text{otherwise},
\end{cases}\\
c_k&\geq P\quad\Longrightarrow\quad\text{freeze }\mathcal{M}_\theta.
\end{aligned}
\end{equation}

The early-stopping rule freezes the uncertainty head after $P$ consecutive
validation evaluations with decreasing PSNR ($\epsilon=0$), while the Gaussian
parameters continue to be optimised. It prevents late-stage uncertainty updates
from adapting to transient photometric residuals after the global scene
structure has largely settled.

\FloatBarrier

\subsection{Implementation Details}
\label{sec:appendix_implementation}

Table~\ref{tab:uncertainty_hparams} lists the default hyper-parameters for
the uncertainty head, opacity modulation, detached soft dropout, and
validation-based early stopping.
Algorithm~\ref{alg:ugod} summarises the full UGOD training and inference
procedure used in all reported experiments.

\begin{table*}[t]
\centering
\footnotesize
\setlength{\tabcolsep}{2pt}
\begin{tabular}{llc}
\toprule
Symbol & Description & Value \\
\midrule
$\kappa$ & modulation slope & $4.0$ \\
$\tau_g$ & modulation threshold & $0.8$ \\
$\beta$ & modulation floor (min.\ opacity multiplier) & $0.70$ \\
$T_g$ & modulation warm-up (iters) & $1200$ \\
\midrule
$\eta$ & dropout scale & $0.08$ \\
$\tau_d$ & dropout threshold & $0.0$ \\
$T_d$ & fixed dropout warm-up (iters) & $1200$ \\
$R$ & dropout ramp duration (iters) & $500$ \\
$T_{\mathrm{conc}}$ & Concrete-mask temperature & $0.1$ \\
$[m_{\min},m_{\max}]$ & keep-mask clamp & $[0.05,\,1.0]$ \\
$c$ & relative-uncertainty clip & $2.0$ \\
\midrule
$\epsilon$ & PSNR-freeze tolerance & $0$ \\
$P$ & PSNR-freeze patience & $2$ \\
\bottomrule
\end{tabular}
\caption{Default hyper-parameters for the uncertainty head, opacity modulation,
detached soft dropout, and early stopping.}
\label{tab:uncertainty_hparams}
\end{table*}

\vspace{-0.5em}
\begin{algorithm}[H]
\footnotesize
\caption{Full UGOD training and inference}
\label{alg:ugod}
\begin{algorithmic}[1]
\REQUIRE Training views $\mathcal{I}$, Gaussians $\mathcal{G}$, budget $T_{\max}$
\ENSURE Optimised $\mathcal{G}$ and uncertainty head $\mathcal{M}_\theta$
\STATE Precompute DPT priors $\mathcal{D}$ on GT training views; initialise $\mathcal{G}$ and $\mathcal{M}_\theta$
\STATE Build pseudo-camera pool (LLFF / $360^\circ$ random poses)
\FOR{$t=1$ \TO $T_{\max}$}
    \STATE Sample a training view; visible set $\mathcal{V}$
    \STATE Form $\mathbf{f}_i$ and predict $u_i=\sigma(\mathcal{M}_\theta(\mathbf{f}_i))$ for $i\in\mathcal{V}$
    \STATE Compute $\alpha_i^{\mathrm{raw}}$ using Eq.~(12) of the main paper
    \IF{$t<T_g$}
        \STATE Render with $\alpha_i^{\mathrm{raw}}$
    \ELSE
        \STATE Normalise $\{u_i\}$ (Eq.~(8) of the main paper); compute $\alpha_i^{\mathrm{gate}}$ (Eq.~(10) of the main paper)
        \IF{$t\geq T_d$}
            \STATE Render with detached soft dropout: $\alpha_i^{\mathrm{gate}}m_i$
        \ELSE
            \STATE Render with $\alpha_i^{\mathrm{gate}}$
        \ENDIF
    \ENDIF
    \STATE Compute photometric loss and $\mathcal{L}_{\mathrm{depth}}$ vs.\ cached prior (Eq.~(17) of the main paper)
    \IF{$500\leq t\leq5500$ and $t$ hits the sampling interval $S$}
        \STATE Sample a pseudo camera; render RGB and depth
        \STATE Run frozen DPT-Hybrid \emph{online} on rendered RGB to obtain $D_{\mathrm{p}}$
        \STATE Add $\alpha(t)\lambda_p\mathcal{L}_{\mathrm{pseudo}}$ (Eq.~(18) of the main paper)
    \ENDIF
    \STATE Form $\mathcal{L}$ using Eq.~(19) of the main paper
    \STATE Update $\mathcal{G}$ and $\mathcal{M}_\theta$ (unless frozen)
    \IF{evaluation step}
        \STATE Freeze $\mathcal{M}_\theta$ if $\Delta_k<\epsilon$ for $P$ consecutive checks
    \ENDIF
\ENDFOR
\STATE \textbf{Inference:} predict $\{u_i\}$ and render with $\alpha_i^{\mathrm{raw}}$
\end{algorithmic}
\end{algorithm}

\subsection{Per-Scene Quantitative Results}
\label{sec:appendix_quantitative}

Tables~\ref{tab:m360_8_rearranged}, \ref{tab:m360_24_rearranged},
and~\ref{tab:llff_3view} report scene-wise PSNR, SSIM, LPIPS, and final
Gaussian counts under the MiP-NeRF~360 8-view, MiP-NeRF~360 24-view, and
LLFF 3-view protocols, respectively.
These tables underlie the averages in Table~1 of the main paper and
expose the scene-dependent quality--compactness trade-offs summarised in the
main text.
Best and second-best values within each scene are in bold and underlined;
our full model is shaded in blue.

\begin{table*}[t]
    \centering
    \caption{Per-scene results on MiP-NeRF~360 (8 views). The full model is
    shaded; best and second-best values are bold and underlined.}
    \label{tab:m360_8_rearranged}
    \begingroup
    \setlength{\tabcolsep}{3pt}
    \renewcommand{\arraystretch}{1.03}
    \footnotesize
    \begin{tabular}{llcccccc}
        \toprule
        \multirow{2}{*}{Scene} & \multirow{2}{*}{Method} & \multirow{2}{*}{Init} & \multicolumn{3}{c}{Metrics} & \multirow{2}{*}{Final Gaussians} & \multirow{2}{*}{Reduction} \\
        \cmidrule(lr){4-6}
        & & & PSNR $\uparrow$ & SSIM $\uparrow$ & LPIPS $\downarrow$ & & vs 3DGS* \\
        \midrule
\multirow{4}{*}{bicycle\_8}
 & 3DGS* & 3 & 11.01 & 0.173 & \best{0.592} & 1{,}460{,}812 & - \\
 & DropGS & 3 & \second{12.24} & \second{0.208} & \second{0.597} & 1{,}000{,}235 & 1.5$\times$ fewer \\
 & FSGS (w/ depth) & 3 & 11.05 & 0.202 & 0.626 & \second{142{,}066} & \second{10.3$\times$ fewer} \\
 \rowcolor{oursrow}
 & Ours (full) & 3 & \best{13.11} & \best{0.284} & 0.643 & \best{59{,}483} & \best{24.6$\times$ fewer} \\
\midrule
\multirow{4}{*}{bonsai\_8}
 & 3DGS* & 11 & 11.9641 & 0.3527 & 0.5958 & 565{,}053 & - \\
 & DropGS & 11 & 12.2382 & 0.3840 & 0.5993 & 338{,}245 & 1.7$\times$ fewer \\
 & FSGS (w/ depth) & 11 & \second{13.6536} & \best{0.4405} & \second{0.5828} & \second{160{,}428} & \second{3.5$\times$ fewer} \\
 \rowcolor{oursrow}
 & Ours (full) & 11 & \best{13.7776} & \second{0.4309} & \best{0.5643} & \best{111{,}546} & \best{5.1$\times$ fewer} \\
\midrule
\multirow{4}{*}{counter\_8}
 & 3DGS* & 51 & 13.9599 & 0.4334 & 0.5710 & 339{,}718 & - \\
 & DropGS & 51 & \best{14.8654} & \best{0.5217} & \best{0.5454} & 343{,}141 & 1.0$\times$ more \\
 & FSGS (w/ depth) & 51 & 13.7273 & 0.5148 & 0.5755 & \best{71{,}677} & \best{4.7$\times$ fewer} \\
 \rowcolor{oursrow}
 & Ours (full) & 51 & \second{14.6130} & \second{0.5185} & \second{0.5565} & \second{77{,}324} & \second{4.4$\times$ fewer} \\
\midrule
\multirow{4}{*}{garden\_8}
 & 3DGS* & 358 & 10.7374 & 0.1449 & \second{0.6499} & 1{,}219{,}741 & - \\
 & DropGS & 358 & 12.6463 & 0.2033 & \best{0.6298} & 922{,}348 & 1.3$\times$ fewer \\
 & FSGS (w/ depth) & 358 & \second{13.4074} & \best{0.2725} & 0.6949 & \best{123{,}253} & \best{9.9$\times$ fewer} \\
 \rowcolor{oursrow}
 & Ours (full) & 358 & \best{14.3317} & \second{0.2592} & 0.6782 & \second{152{,}968} & \second{8.0$\times$ fewer} \\
\midrule
\multirow{4}{*}{kitchen\_8}
 & 3DGS* & 22 & 12.9750 & 0.3761 & 0.6391 & 490{,}670 & - \\
 & DropGS & 22 & 12.6709 & 0.3938 & 0.6349 & 803{,}427 & 1.6$\times$ more \\
 & FSGS (w/ depth) & 22 & \second{13.0945} & \best{0.4510} & \best{0.6153} & \best{30{,}328} & \best{16.2$\times$ fewer} \\
 \rowcolor{oursrow}
 & Ours (full) & 22 & \best{15.0121} & \second{0.4271} & \second{0.6292} & \second{40{,}950} & \second{12.0$\times$ fewer} \\
\midrule
\multirow{4}{*}{stump\_8}
 & 3DGS* & 14 & 11.1085 & 0.1684 & \second{0.6094} & 1{,}110{,}366 & - \\
 & DropGS & 14 & 11.6992 & 0.2049 & \best{0.6089} & 889{,}400 & 1.2$\times$ fewer \\
 & FSGS (w/ depth) & 14 & \second{13.2528} & \second{0.2720} & 0.6237 & \second{264{,}785} & \second{4.2$\times$ fewer} \\
 \rowcolor{oursrow}
 & Ours (full) & 14 & \best{14.0165} & \best{0.2897} & 0.6363 & \best{131{,}701} & \best{8.4$\times$ fewer} \\
        \bottomrule
    \end{tabular}
    \endgroup
\end{table*}

\begin{table*}[t]
    \centering
    \caption{Per-scene results on MiP-NeRF~360 (24 views). The full model is
    shaded; best and second-best values are bold and underlined.}
    \label{tab:m360_24_rearranged}
    \begingroup
    \setlength{\tabcolsep}{3.5pt}
    \renewcommand{\arraystretch}{1.05}
    \footnotesize
    \begin{tabular}{llcccccc}
        \toprule
        \multirow{2}{*}{Scene} & \multirow{2}{*}{Method} & \multirow{2}{*}{Init} & \multicolumn{3}{c}{Metrics} & \multirow{2}{*}{Final Gaussians} & \multirow{2}{*}{Reduction} \\
        \cmidrule(lr){4-6}
        & & & PSNR $\uparrow$ & SSIM $\uparrow$ & LPIPS $\downarrow$ & & vs 3DGS* \\
        \midrule
\multirow{4}{*}{bicycle\_24}
 & 3DGS* & 21 & 15.3990 & 0.2815 & \best{0.5713} & 1{,}292{,}646 & - \\
 & DropGS & 21 & \second{15.8850} & 0.3172 & \second{0.5894} & 937{,}115 & 1.4$\times$ fewer \\
 & FSGS (w/ depth) & 21 & 15.8226 & \second{0.3482} & 0.6068 & \second{155{,}215} & \second{8.3$\times$ fewer} \\
 \rowcolor{oursrow}
 & Ours (full) & 21 & \best{16.8980} & \best{0.3602} & 0.5954 & \best{141{,}530} & \best{9.1$\times$ fewer} \\
\midrule
\multirow{4}{*}{bonsai\_24}
 & 3DGS* & 5764 & \second{22.5979} & \second{0.8081} & \best{0.2986} & 686{,}826 & - \\
 & DropGS & 5764 & \best{22.6805} & \best{0.8113} & \second{0.3001} & 600{,}128 & 1.1$\times$ fewer \\
 & FSGS (w/ depth) & 5764 & 22.1688 & 0.7529 & 0.3903 & \second{176{,}229} & \second{3.9$\times$ fewer} \\
 \rowcolor{oursrow}
 & Ours (full) & 5764 & 20.1803 & 0.7038 & 0.4269 & \best{154{,}205} & \best{4.5$\times$ fewer} \\
\midrule
\multirow{4}{*}{counter\_24}
 & 3DGS* & 3572 & 20.7878 & \best{0.7505} & \best{0.3479} & 620{,}747 & - \\
 & DropGS & 3572 & \second{20.9882} & \second{0.7470} & \second{0.3716} & 506{,}551 & 1.2$\times$ fewer \\
 & FSGS (w/ depth) & 3572 & 20.4888 & 0.6933 & 0.4546 & \second{107{,}532} & \second{5.8$\times$ fewer} \\
 \rowcolor{oursrow}
 & Ours (full) & 3572 & \best{21.6592} & 0.7263 & 0.4288 & \best{93{,}907} & \best{6.6$\times$ fewer} \\
\midrule
\multirow{4}{*}{stump\_24}
 & 3DGS* & 1013 & 14.3681 & 0.3311 & \second{0.5889} & 1{,}223{,}612 & - \\
 & DropGS & 1013 & 15.2976 & 0.3613 & \best{0.5823} & 952{,}989 & 1.3$\times$ fewer \\
 & FSGS (w/ depth) & 1013 & \best{17.0636} & \second{0.4036} & 0.6088 & \second{149{,}268} & \second{8.2$\times$ fewer} \\
 \rowcolor{oursrow}
 & Ours (full) & 1013 & \second{16.7614} & \best{0.4100} & 0.6104 & \best{142{,}794} & \best{8.6$\times$ fewer} \\
\midrule
\multirow{4}{*}{garden\_24}
 & 3DGS* & 1343 & 19.6501 & \best{0.5328} & \best{0.3816} & 1{,}737{,}612 & - \\
 & DropGS & 1343 & \best{19.9231} & \second{0.4966} & \second{0.4473} & 1{,}154{,}896 & 1.5$\times$ fewer \\
 & FSGS (w/ depth) & 1343 & 19.6418 & 0.4337 & 0.5468 & \second{218{,}774} & \second{7.9$\times$ fewer} \\
 \rowcolor{oursrow}
 & Ours (full) & 1343 & \second{19.7033} & 0.4103 & 0.5647 & \best{154{,}324} & \best{11.3$\times$ fewer} \\
\midrule
\multirow{4}{*}{kitchen\_24}
 & 3DGS* & 268 & \second{19.2408} & \best{0.6644} & \best{0.3848} & 672{,}266 & - \\
 & DropGS & 268 & 19.1724 & \second{0.6443} & \second{0.4212} & 502{,}238 & 1.3$\times$ fewer \\
 & FSGS (w/ depth) & 268 & 17.5143 & 0.5576 & 0.5397 & \best{82{,}761} & \best{8.1$\times$ fewer} \\
 \rowcolor{oursrow}
 & Ours (full) & 268 & \best{19.3180} & 0.6087 & 0.4956 & \second{90{,}226} & \second{7.5$\times$ fewer} \\
        \bottomrule
    \end{tabular}
    \endgroup
\end{table*}

\begin{table*}[t]
\centering
\caption{Per-scene results on LLFF (3 views). The full model is shaded; best
and second-best values are bold and underlined.}
\label{tab:llff_3view}
\begingroup
\setlength{\tabcolsep}{3pt}
\renewcommand{\arraystretch}{1.03}
\footnotesize
\begin{tabular}{llrccccc}
\toprule
Scene & Method & Init & \multicolumn{3}{c}{Metrics} & Final & Compression \\
\cmidrule(lr){4-6}
 & & Points & PSNR$\uparrow$ & SSIM$\uparrow$ & LPIPS$\downarrow$ & Gaussians & vs 3DGS* \\
\midrule
\multirow{4}{*}{Fern}
 & 3DGS* & 1707 & 13.99 & 0.444 & 0.421 & 325{,}283 & -- \\
 & DropGS & 1707 & 13.69 & 0.448 & 0.425 & 245{,}406 & 1.3$\times$ fewer \\
 & FSGS (w/ depth) & 1707 & \second{20.49} & \second{0.665} & \second{0.243} & \second{117{,}432} & \second{2.8$\times$ fewer} \\
 \rowcolor{oursrow}
 & Ours (full) & 1707 & \best{20.74} & \best{0.680} & \best{0.231} & \best{93{,}766} & \best{3.5$\times$ fewer} \\
\midrule
\multirow{4}{*}{Flower}
 & 3DGS* & 1804 & 16.81 & 0.479 & 0.371 & 233{,}646 & -- \\
 & DropGS & 1804 & 16.94 & 0.479 & 0.375 & \second{168{,}128} & \second{1.4$\times$ fewer} \\
 & FSGS (w/ depth) & 1804 & \second{19.79} & \second{0.599} & \second{0.283} & 244{,}626 & 1.0$\times$ more \\
 \rowcolor{oursrow}
 & Ours (full) & 1804 & \best{19.96} & \best{0.616} & \best{0.264} & \best{93{,}361} & \best{2.5$\times$ fewer} \\
\midrule
\multirow{4}{*}{Fortress}
 & 3DGS* & 2744 & 18.74 & 0.605 & 0.251 & 153{,}393 & -- \\
 & DropGS & 2744 & 17.26 & 0.600 & 0.253 & 154{,}500 & 1.0$\times$ more \\
 & FSGS (w/ depth) & 2744 & \second{21.96} & \second{0.687} & \second{0.195} & \second{61{,}502} & \second{2.5$\times$ fewer} \\
 \rowcolor{oursrow}
 & Ours (full) & 2744 & \best{22.98} & \best{0.697} & \best{0.190} & \best{54{,}545} & \best{2.8$\times$ fewer} \\
\midrule
\multirow{4}{*}{Horns}
 & 3DGS* & 2643 & 15.25 & 0.470 & 0.390 & 190{,}676 & -- \\
 & DropGS & 2643 & 15.23 & 0.489 & 0.386 & 157{,}449 & 1.2$\times$ fewer \\
 & FSGS (w/ depth) & 2643 & \second{19.07} & \second{0.644} & \second{0.276} & \second{99{,}171} & \second{1.9$\times$ fewer} \\
 \rowcolor{oursrow}
 & Ours (full) & 2643 & \best{19.12} & \best{0.661} & \best{0.273} & \best{71{,}057} & \best{2.7$\times$ fewer} \\
\midrule
\multirow{4}{*}{Leaves}
 & 3DGS* & 726 & 12.23 & 0.280 & 0.441 & 565{,}899 & -- \\
 & DropGS & 726 & 12.77 & 0.306 & 0.412 & 459{,}942 & 1.2$\times$ fewer \\
 & FSGS (w/ depth) & 726 & \second{16.10} & \second{0.522} & \second{0.257} & \second{403{,}301} & \second{1.4$\times$ fewer} \\
 \rowcolor{oursrow}
 & Ours (full) & 726 & \best{16.64} & \best{0.566} & \best{0.246} & \best{276{,}785} & \best{2.0$\times$ fewer} \\
\midrule
\multirow{4}{*}{Orchids}
 & 3DGS* & 867 & 14.18 & 0.377 & 0.360 & 204{,}548 & -- \\
 & DropGS & 867 & 14.49 & 0.405 & 0.336 & 188{,}259 & 1.1$\times$ fewer \\
 & FSGS (w/ depth) & 867 & \best{15.28} & \best{0.437} & \best{0.311} & \second{97{,}790} & \second{2.1$\times$ fewer} \\
 \rowcolor{oursrow}
 & Ours (full) & 867 & \second{15.21} & \best{0.437} & \second{0.313} & \best{89{,}302} & \best{2.3$\times$ fewer} \\
\midrule
\multirow{4}{*}{Room}
 & 3DGS* & 1026 & 11.38 & 0.410 & 0.559 & 107{,}004 & -- \\
 & DropGS & 1026 & 12.13 & 0.463 & 0.539 & 99{,}216 & 1.1$\times$ fewer \\
 & FSGS (w/ depth) & 1026 & \second{20.63} & \best{0.807} & \best{0.195} & \best{34{,}386} & \best{3.1$\times$ fewer} \\
 \rowcolor{oursrow}
 & Ours (full) & 1026 & \best{20.66} & \second{0.802} & \second{0.197} & \second{37{,}146} & \second{2.9$\times$ fewer} \\
\bottomrule
\end{tabular}
\endgroup
\end{table*}

\subsection{Component Ablation Details}
\label{sec:appendix_ablation}

Tables~\ref{tab:ablation_m360_8}, \ref{tab:ablation_m360_24},
and~\ref{tab:ablation_llff_3} provide the per-scene component ablation
underlying the averages in Table~2 of the main paper.
We compare intermediate variants modulation only, with/without soft dropout,
and the full configuration within each scene.
Best and second-best values within each scene are in bold and underlined;
the full model is shaded in blue.

\begin{table*}[!b]
\centering
\caption{Per-scene component ablation on MiP-NeRF~360 (8 views). Best and
second-best values are bold and underlined.}
\label{tab:ablation_m360_8}
\begingroup
\setlength{\tabcolsep}{3pt}
\renewcommand{\arraystretch}{1.03}
\footnotesize
\begin{tabular}{llccccc}
\toprule
Scene & Variant & PSNR$\uparrow$ & SSIM$\uparrow$ & LPIPS$\downarrow$ & Final Gaussians & vs 3DGS* \\
\midrule
 & Ours (w/o soft dropout) & \second{11.24} & 0.239 & \second{0.688} & \second{24{,}313} & \second{60.1$\times$ fewer} \\
 & Ours (w/ soft dropout) & 10.56 & \second{0.246} & 0.689 & \best{14{,}400} & \best{101.4$\times$ fewer} \\
 \rowcolor{oursrow}
\multirow{-3}{*}{bicycle} & Ours (full) & \best{13.11} & \best{0.284} & \best{0.643} & 59{,}483 & 24.6$\times$ fewer \\
\cmidrule{1-7}
 & Ours (w/o soft dropout) & \best{13.9212} & \best{0.4432} & \second{0.5847} & \second{97{,}801} & \second{5.8$\times$ fewer} \\
 & Ours (w/ soft dropout) & 12.6963 & \second{0.4385} & 0.6108 & \best{52{,}688} & \best{10.7$\times$ fewer} \\
 \rowcolor{oursrow}
\multirow{-3}{*}{bonsai} & Ours (full) & \second{13.7776} & 0.4309 & \best{0.5643} & 111{,}546 & 5.1$\times$ fewer \\
\cmidrule{1-7}
 & Ours (w/o soft dropout) & \second{14.0737} & \second{0.4916} & \second{0.5658} & \second{79{,}049} & \second{4.3$\times$ fewer} \\
 & Ours (w/ soft dropout) & 13.5822 & 0.4629 & 0.5947 & 85{,}530 & 4.0$\times$ fewer \\
 \rowcolor{oursrow}
\multirow{-3}{*}{counter} & Ours (full) & \best{14.6130} & \best{0.5185} & \best{0.5565} & \best{77{,}324} & \best{4.4$\times$ fewer} \\
\cmidrule{1-7}
 & Ours (w/o soft dropout) & \second{13.4690} & \second{0.2283} & \best{0.6510} & 239{,}534 & 5.1$\times$ fewer \\
 & Ours (w/ soft dropout) & 12.6834 & 0.2135 & \second{0.6752} & \second{226{,}193} & \second{5.4$\times$ fewer} \\
 \rowcolor{oursrow}
\multirow{-3}{*}{garden} & Ours (full) & \best{14.3317} & \best{0.2592} & 0.6782 & \best{152{,}968} & \best{8.0$\times$ fewer} \\
\cmidrule{1-7}
 & Ours (w/o soft dropout) & \second{14.4344} & \second{0.4102} & \best{0.6132} & 56{,}932 & 8.6$\times$ fewer \\
 & Ours (w/ soft dropout) & 12.1802 & 0.3808 & 0.6454 & \second{56{,}480} & \second{8.7$\times$ fewer} \\
 \rowcolor{oursrow}
\multirow{-3}{*}{kitchen} & Ours (full) & \best{15.0121} & \best{0.4271} & \second{0.6292} & \best{40{,}950} & \best{12.0$\times$ fewer} \\
\cmidrule{1-7}
 & Ours (w/o soft dropout) & \second{13.8207} & \second{0.2626} & \second{0.6225} & \best{66{,}723} & \best{16.6$\times$ fewer} \\
 & Ours (w/ soft dropout) & 13.7282 & 0.2418 & \best{0.6072} & \second{92{,}331} & \second{12.0$\times$ fewer} \\
 \rowcolor{oursrow}
\multirow{-3}{*}{stump} & Ours (full) & \best{14.0165} & \best{0.2897} & 0.6363 & 131{,}701 & 8.4$\times$ fewer \\
\bottomrule
\end{tabular}
\endgroup
\end{table*}

\begin{table*}[t]
\centering
\caption{Per-scene component ablation on MiP-NeRF~360 (24 views). Best and
second-best values are bold and underlined.}
\label{tab:ablation_m360_24}
\begingroup
\setlength{\tabcolsep}{3pt}
\renewcommand{\arraystretch}{1.03}
\footnotesize
\begin{tabular}{llccccc}
\toprule
Scene & Variant & PSNR$\uparrow$ & SSIM$\uparrow$ & LPIPS$\downarrow$ & Final Gaussians & vs 3DGS* \\
\midrule
\multirow{3}{*}{bicycle}
 & Ours (w/o soft dropout) & \second{16.6135} & \second{0.3429} & \second{0.6026} & \second{84{,}962} & \second{15.2$\times$ fewer} \\
 & Ours (w/ soft dropout) & 15.7267 & 0.3208 & 0.6139 & \best{76{,}494} & \best{16.9$\times$ fewer} \\
 \rowcolor{oursrow}
 & Ours (full) & \best{16.8980} & \best{0.3602} & \best{0.5954} & 141{,}530 & 9.1$\times$ fewer \\
\cmidrule{1-7}
\multirow{3}{*}{bonsai}
 & Ours (w/o soft dropout) & \best{22.1005} & \best{0.7864} & \best{0.3320} & \second{337{,}290} & \second{2.0$\times$ fewer} \\
 & Ours (w/ soft dropout) & \second{21.7632} & \second{0.7790} & \second{0.3339} & 344{,}817 & 2.0$\times$ fewer \\
 \rowcolor{oursrow}
 & Ours (full) & 20.1803 & 0.7038 & 0.4269 & \best{154{,}205} & \best{4.5$\times$ fewer} \\
\cmidrule{1-7}
\multirow{3}{*}{counter}
 & Ours (w/o soft dropout) & \second{21.6241} & \best{0.7449} & \best{0.3738} & 218{,}612 & 2.8$\times$ fewer \\
 & Ours (w/ soft dropout) & 20.8360 & \second{0.7349} & \second{0.3767} & \second{215{,}300} & \second{2.9$\times$ fewer} \\
 \rowcolor{oursrow}
 & Ours (full) & \best{21.6592} & 0.7263 & 0.4288 & \best{93{,}907} & \best{6.6$\times$ fewer} \\
\cmidrule{1-7}
\multirow{3}{*}{stump}
 & Ours (w/o soft dropout) & 16.2540 & \second{0.3867} & \second{0.5827} & 273{,}927 & 4.5$\times$ fewer \\
 & Ours (w/ soft dropout) & \second{16.4707} & 0.3691 & \best{0.5824} & \second{254{,}156} & \second{4.8$\times$ fewer} \\
 \rowcolor{oursrow}
 & Ours (full) & \best{16.7614} & \best{0.4100} & 0.6104 & \best{142{,}794} & \best{8.6$\times$ fewer} \\
\cmidrule{1-7}
\multirow{3}{*}{garden}
 & Ours (w/o soft dropout) & \second{19.1356} & \second{0.4345} & \second{0.5028} & \second{359{,}334} & \second{4.8$\times$ fewer} \\
 & Ours (w/ soft dropout) & 18.9934 & \best{0.4352} & \best{0.5019} & 385{,}015 & 4.5$\times$ fewer \\
 \rowcolor{oursrow}
 & Ours (full) & \best{19.7033} & 0.4103 & 0.5647 & \best{154{,}324} & \best{11.3$\times$ fewer} \\
\cmidrule{1-7}
\multirow{3}{*}{kitchen}
 & Ours (w/o soft dropout) & \second{18.7236} & \second{0.6172} & \second{0.4525} & \second{188{,}888} & \second{3.6$\times$ fewer} \\
 & Ours (w/ soft dropout) & 18.6843 & \best{0.6184} & \best{0.4470} & 190{,}147 & 3.5$\times$ fewer \\
 \rowcolor{oursrow}
 & Ours (full) & \best{19.3180} & 0.6087 & 0.4956 & \best{90{,}226} & \best{7.5$\times$ fewer} \\
\bottomrule
\end{tabular}
\endgroup
\end{table*}

\begin{table*}[t]
\centering
\caption{Per-scene component ablation on LLFF (3 views). Best and second-best
values are bold and underlined.}
\label{tab:ablation_llff_3}
\begingroup
\setlength{\tabcolsep}{3pt}
\renewcommand{\arraystretch}{1.03}
\footnotesize
\begin{tabular}{llccccc}
\toprule
Scene & Variant & PSNR$\uparrow$ & SSIM$\uparrow$ & LPIPS$\downarrow$ & Final Gaussians & vs 3DGS* \\
\midrule
\multirow{3}{*}{Fern}
 & Ours & \second{20.09} & \second{0.649} & \second{0.248} & \second{198{,}554} & \second{1.6$\times$ fewer} \\
 & Ours (w/o soft dropout) & 19.06 & 0.605 & 0.277 & 204{,}963 & 1.6$\times$ fewer \\
 \rowcolor{oursrow}
 & Ours (full) & \best{20.74} & \best{0.680} & \best{0.231} & \best{93{,}766} & \best{3.5$\times$ fewer} \\
\cmidrule{1-7}
\multirow{3}{*}{Flower}
 & Ours & \second{19.85} & \best{0.619} & \best{0.251} & 193{,}944 & 1.2$\times$ fewer \\
 & Ours (w/o soft dropout) & 18.97 & 0.591 & 0.271 & \second{192{,}774} & \second{1.2$\times$ fewer} \\
 \rowcolor{oursrow}
 & Ours (full) & \best{19.96} & \second{0.616} & \second{0.264} & \best{93{,}361} & \best{2.5$\times$ fewer} \\
\cmidrule{1-7}
\multirow{3}{*}{Fortress}
 & Ours & \second{22.37} & \best{0.697} & \second{0.195} & 113{,}292 & \second{1.4$\times$ fewer} \\
 & Ours (w/o soft dropout) & 21.43 & 0.644 & 0.222 & \second{112{,}321} & 1.4$\times$ fewer \\
 \rowcolor{oursrow}
 & Ours (full) & \best{22.98} & \best{0.697} & \best{0.190} & \best{54{,}545} & \best{2.8$\times$ fewer} \\
\cmidrule{1-7}
\multirow{3}{*}{Horns}
 & Ours & \second{17.59} & \second{0.585} & \second{0.309} & \second{135{,}013} & \second{1.4$\times$ fewer} \\
 & Ours (w/o soft dropout) & 17.30 & 0.541 & 0.339 & 136{,}915 & 1.4$\times$ fewer \\
 \rowcolor{oursrow}
 & Ours (full) & \best{19.12} & \best{0.661} & \best{0.273} & \best{71{,}057} & \best{2.7$\times$ fewer} \\
\cmidrule{1-7}
\multirow{3}{*}{Leaves}
 & Ours & \second{15.31} & \second{0.491} & \second{0.289} & \second{382{,}086} & \second{1.5$\times$ fewer} \\
 & Ours (w/o soft dropout) & 15.09 & 0.464 & 0.310 & 387{,}407 & 1.5$\times$ fewer \\
 \rowcolor{oursrow}
 & Ours (full) & \best{16.64} & \best{0.566} & \best{0.246} & \best{276{,}785} & \best{2.0$\times$ fewer} \\
\cmidrule{1-7}
\multirow{3}{*}{Orchids}
 & Ours & \second{14.76} & \second{0.412} & \second{0.327} & \second{175{,}424} & \second{1.2$\times$ fewer} \\
 & Ours (w/o soft dropout) & 14.44 & 0.394 & 0.344 & 177{,}430 & 1.2$\times$ fewer \\
 \rowcolor{oursrow}
 & Ours (full) & \best{15.21} & \best{0.437} & \best{0.313} & \best{89{,}302} & \best{2.3$\times$ fewer} \\
\cmidrule{1-7}
\multirow{3}{*}{Room}
 & Ours & 17.70 & 0.739 & 0.272 & \second{68{,}023} & \second{1.6$\times$ fewer} \\
 & Ours (w/o soft dropout) & \second{19.38} & \second{0.751} & \second{0.251} & 72{,}410 & 1.5$\times$ fewer \\
 \rowcolor{oursrow}
 & Ours (full) & \best{20.66} & \best{0.802} & \best{0.197} & \best{37{,}146} & \best{2.9$\times$ fewer} \\
\bottomrule
\end{tabular}
\endgroup
\end{table*}

\end{document}